\documentclass[lettersize,journal]{IEEEtran}
\usepackage{amsmath,amsfonts}
\usepackage{algorithm}
\usepackage{algpseudocode}
\usepackage{array}
\usepackage[caption=false,font=normalsize,labelfont=sf,textfont=sf]{subfig}
\usepackage{textcomp}
\usepackage{stfloats}
\usepackage{url}
\usepackage{xcolor}
\usepackage{verbatim}
\usepackage{graphicx}
\usepackage{cite}
\usepackage{multirow}
\usepackage{makecell}
\usepackage{booktabs}
\usepackage{longtable}
\usepackage{pifont} 
\usepackage[T1]{fontenc}
\usepackage{aecompl}
\usepackage{xcolor}
\usepackage[hidelinks]{hyperref}
\usepackage{pdfpages}

\def\ie{\emph{i.e.}}
\def\eg{\emph{e.g.}}

\begin{document}

\title{OPA-Pack: Object-Property-Aware\\ Robotic Bin Packing}

\author{ 
Jia-Hui Pan, Yeok Tatt Cheah, Zhengzhe Liu, Ka-Hei Hui, Xiaojie Gao, \\ Pheng-Ann Heng, Yun-Hui Liu, and Chi-Wing Fu
\thanks{}
\thanks{}
\thanks{
%
This study was supported by the InnoHK initiative of the Innovation and Technology Commission of the Hong Kong Special Administrative Region Government via the Hong Kong Centre for Logistics Robotics.
This work was also supported by NSFC (Category C) fund code 62506149.
}
\thanks{Jia-Hui Pan, Xiaojie Gao, Pheng-Ann Heng, and Chi-Wing Fu are with the Department of Computer Science and Engineering, the Chinese University of Hong Kong (e-mail: \{jhpan21, xjgao, pheng, cwfu\}@cse.cuhk.edu.hk);
Yeok Tatt Cheah is with the Hong Kong Centre for Logistics Robotics (e-mail:  ytcheah@hkclr.hk);
Zhengzhe Liu is with the School of Data Science, Lingnan University (e-mail: zhengzheliu@ln.edu.hk);
Ka-Hei Hui is with Autodesk AI Lab (Canada) (e-mail:  ka.hei.hui@autodesk.com);
Yun-Hui Liu is with the Department of Mechanical and Automation Engineering, the Chinese University of Hong Kong (e-mail: yhliu@mae.cuhk.edu.hk).
}
\thanks{We thank Ziheng Kang for scanning and processing objects for real-world experiments, Jichun Yang for maintaining the packing system, and Xieyuanli Chen, Tianyu Huang, Kai Chen, Wei Chen, Runsong Zhu, and Zhiliang Zeng for suggestions on technical design and academic writing.}
}


\markboth{}%
{OPA-Pack: Object-Property-Aware\\Robotic Bin Packing}

\maketitle

\begin{abstract}
Robotic bin packing is a fundamental manipulation problem with broad applications in e-commerce and warehouse automation.
However, existing methods primarily focus on object geometry to optimize packing compactness without explicitly modeling object properties such as fragility, edibility, and chemical compatibility that humans typically consider during packing.
This paper presents OPA-Pack (Object-Property-Aware Packing), a framework that incorporates object-property information into robotic bin-packing planning.
The framework includes an object property recognition scheme based on retrieval-augmented generation and chain-of-thought reasoning to establish a dataset with object property annotations for 1,032 everyday objects.
The framework further incorporates OPA-Net,
which encodes candidate-object properties via a property embedding layer and represents the container state using fragility and avoidance heightmaps.
We design a reward function and adopt a deep Q-learning scheme to train OPA-Net.
Experimental results show that OPA-Pack improves avoidance accuracy from 52.3\% to 95.0\%, reduces load on fragile objects by 29.4\%, and achieves a packing compactness of 0.413, compared with 0.420 for the baseline.
In addition, we evaluate OPA-Pack on a physical platform using 70 packing sequences sampled from 100 scanned real-world objects.
Results show that OPA-Pack
outperforms the baseline and other packing heuristics in reducing the number of closely packed incompatible object pairs and the number of compressed fragile objects while preserving packing compactness.
Project page: \href{https://kwpoon.github.io/OPA-Pack/}{https://kwpoon.github.io/OPA-Pack/}.
%
\end{abstract}

\begin{IEEEkeywords}
Robotic bin packing,
manipulation planning, 
object property
\end{IEEEkeywords}


\begin{center}
    \normalsize \textbf{\IEEEPARstart{}{List of Notation}}
\end{center}

\hspace*{-13pt}
\begin{tabular}{@{}p{0.22\columnwidth}@{\hspace{10pt}}p{\dimexpr0.78\columnwidth-2pt\relax}@{}}
    $\mathbf{s}_{t}$ & Packing state at packing step $t$ \\
    $\mathbf{a}_{t}$ & Packing action at packing step $t$ \\
    $R(\mathbf{s}_{t}, \mathbf{a}_{t})$ & Reward function \\
\end{tabular}

\hspace*{-13pt}
\begin{tabular}{@{}p{0.22\columnwidth}@{\hspace{10pt}}p{\dimexpr0.78\columnwidth-2pt\relax}@{}}
    $Q(\mathbf{s}_{t}, \mathbf{a}_{t})$ & Q-value function \\
    $\operatorname{obs}(\cdot)$ & Observation of the scene \\
    $\widetilde{Q}(\operatorname{obs}(\mathbf{s}_{t}), \mathbf{a}_{t})$
        & Predicted Q-value function \\
    $\widetilde{V}(\operatorname{obs}(\mathbf{s}_{t}))$
        & Predicted state-value function \\
    $\widetilde{A}(\operatorname{obs}(\mathbf{s}_{t}), \mathbf{a}_{t})$
        & Predicted advantage function \\
    $b$ & A candidate object \\
    $\mathbf{f}_{\mathrm{pcl}}^{b}$ & Point cloud feature for object $b$ \\
    $\mathbf{f}_{\mathrm{pro}}^{b}$ & Property feature for object $b$ \\
    $\mathbf{H}_{\mathrm{occ}}$ & Occupancy heightmap \\
    $\mathbf{H}_{\mathrm{frg}}$ & Fragility heightmap \\
    $\mathbf{H}_{\mathrm{avd}}^b$ & Avoidance heightmap for object $b$ \\
    $\mathbf{f}_{\mathrm{occ}}$ & Occupancy heightmap feature \\
    $\mathbf{f}_{\mathrm{frg}}$ & Fragility heightmap feature \\
    $\mathbf{f}_{\mathrm{avd}}^b$ & Avoidance heightmap feature for object $b$ \\
    $\mathbf{f}_{\mathrm{obj}}^b$ & Feature for object selection \\
    $\widetilde{V}_{\mathrm{obj}}$ & Predicted state value for object selection \\
    $\widetilde{A}_{\mathrm{obj}}^b$ & Predicted advantage value for object selection \\
    $\mathbf{f}_{\mathrm{pos}}^{b}$ & Pose feature for object $b$ \\
    $\mathbf{f}_{\mathrm{place}}$ & Feature for placement selection \\
    $\widetilde{V}_{\mathrm{place}}$ & Predicted state value for placement selection \\
    $\widetilde{A}_{\mathrm{place}}$ & Predicted advantage value for placement selection \\
    $C$ & Packing compactness \\
    $R_{\mathrm{frg}}$ & Fragility reward term \\
    $R_{\mathrm{avd}}$ & Avoidance reward term \\
    $R_{\mathrm{total}}$ & Total reward \\
\end{tabular}

\section{Introduction}
\IEEEPARstart{R}{obotic} bin packing involves arranging objects into a container using a robotic manipulator.
It is a key capability for packaging systems in e-commerce and warehouse logistics, where automation can reduce labor costs, increase throughput, and improve workflow stability.

Prior studies have formulated robotic bin packing as a geometric optimization problem~\cite{zhao2023learning, ma2018packing, yang2023heuristics, jia2022robot, puche2022online, huang2022planning, hu2020tap, wang2019stable, pan2023sdf, pan2024ppn, wang2021dense}.
These methods typically optimize object placements based on geometric information to improve space utilization.
Several approaches have been proposed to improve packing compactness:
(i) optimization-based methods~\cite{ramos2016container, karabulut2004hybrid, kang2012hybrid,crainic2008extreme, liu2015hape3d, lamas2022voxel, jiang2012learning}, which search for compact packing solutions through offline sampling and evaluation, 
(ii) packing heuristics~\cite{karabulut2004hybrid, ramos2016container, wang2019stable, wang2021dense, pan2023sdf}, which use objective functions to evaluate candidate objects and placement options online, and (iii) reinforcement learning~\cite{hu2020tap,zhao2021online,verma2020generalized,zhang2021attend2pack,yang2023heuristics,pan2023adjustable, huang2022planning, zhao2023learning, yang2021packerbot}, which learns sequential packing policies to improve long-term packing compactness.
Several studies~\cite{moon2014container,de2013two} aim to reduce concentrated loading within the container by controlling the placement and distribution of heavy items.
However, broader physical or semantic properties and the incompatibilities among objects remain underexplored.

In practical scenarios, bin packing involves constraints beyond geometric optimization.
Human packing decisions typically account for object properties.
Representative constraints include preventing heavy objects from being placed above fragile objects, separating sharp objects from soft objects, and isolating food products from household chemicals to avoid cross-contamination.
Most existing robotic bin-packing methods overlook these physical and semantic object characteristics, limiting their practical applicability.
Recent advancements in artificial intelligence have facilitated the integration of scene and object semantics into robot manipulation~\cite{firoozi2023foundation, li2024manipllm, ni2024generate, huang2023voxposer, yang2025octopus, huang2024rekep, di2024keypoint}.
To this end, this work explores the integration of object properties into the planning of robotic bin packing.

Incorporating object properties into bin-packing models presents two practical challenges: object property acquisition and property-aware packing planning.
First, property acquisition for everyday objects requires extensive human annotations and external knowledge.
Recent vision-language foundation models, trained on large-scale multimodal data and diverse tasks~\cite{bommasani2021opportunities,achiam2023gpt}, have shown the potential to associate visual observations with object-level semantic and physical attributes. 
These models enable object-property estimation from visual observations for subsequent robotic perception and planning~\cite{xu2024manifoundation}.
Second, property-aware planning requires not only an optimization objective that balances geometric compactness with property-related constraints, but also a multi-step planning process to optimize the packing decisions.

In this work, we present \textbf{OPA-Pack}, an \textbf{O}bject-\textbf{P}roperty-\textbf{A}ware \textbf{Pack}ing framework for everyday objects.
To the best of our knowledge, OPA-Pack is the first robotic packing framework to explicitly integrate object-property information with geometric representations for sequential packing planning.
To address the aforementioned challenges, OPA-Pack consists of two stages: (i) object property recognition and (ii) packing policy learning.

In the first stage, we propose an object property recognition scheme to predict physical attributes (\eg, softness, sharpness, fragility, and density) and semantic properties (\eg, edible items, household chemicals, and medicines).
The property recognition is enhanced using retrieval-augmented generation~\cite{zhao2024retrieval,lewis2020retrieval}, which retrieves similar examples from an auxiliary database to provide additional context for object-property prediction.
For object properties that depend on material composition, such as fragility and density, chain-of-thought reasoning~\cite{wei2022chain,feng2024towards,zhangautomatic,chu2023survey} is employed to estimate the object's inner and outer materials, which are subsequently combined with prior material knowledge to guide property recognition.
The predicted object properties are further utilized to infer avoidance relations between objects, thereby identifying object pairs that should not be placed in close proximity, such as household chemicals and edible items, or sharp objects and soft objects.
An OPA dataset containing 1,032 everyday objects with property annotations is constructed to facilitate property-aware packing.

In the second stage, we develop \textbf{OPA-Net} (\textbf{O}bject-\textbf{P}roperty-\textbf{A}ware packing \textbf{Net}work).
OPA-Net is designed to account for three objectives during packing: 
(i) spatial separation of incompatible object pairs,
(ii) load reduction on fragile objects, and 
(iii) enhancement of packing compactness.
Following \cite{zhao2023learning}, OPA-Net is trained with deep Q-learning using a reward function designed to optimize the three objectives jointly.
The network uses dedicated encoders to capture object geometry, object properties, and container-state information derived from the locations of previously packed objects.
We develop an object Q-predictor to select a candidate object and a placement Q-predictor to determine its placement pose.

This work incorporates object properties into packing policy learning, with the main contributions summarized as follows:
\begin{itemize}
    \item We propose OPA-Pack, a robotic bin-packing framework that incorporates object properties into packing planning;
    \item OPA-Net is designed to separate incompatible object pairs and reduce load on fragile objects while preserving packing compactness;
    \item We introduce an object property recognition scheme based on retrieval-augmented generation and chain-of-thought reasoning; 
    \item We construct a dataset of 1,032 everyday objects with property annotations to support property-aware packing;
    \item Simulation experiments show that OPA-Pack improves avoidance accuracy from 52.3\% to 95.0\%, reduces load on fragile objects by 29.4\%, and achieves a packing compactness of 0.413, while the baseline achieves 0.420;
    \item We collect and annotate 100 scanned real-world objects with property labels and establish a physical platform, showing that OPA-Pack outperforms the baseline and other packing heuristics in real-world settings.
\end{itemize}

\section{Related Work}
Bin packing shares conceptual similarities with the knapsack problem~\cite{ross1989stochastic,chu1998genetic,horowitz1974computing,martinez2013exact,casazza2016exactly}, as both involve item allocation with limited capacity.
The knapsack problem concerns selecting items subject to capacity constraints.
While some formulations~\cite{martinez2013exact,casazza2016exactly} incorporate item attributes such as weight, volume, or fragility, they typically treat objects as abstract entities and do not address the three-dimensional geometric placement constraints required in robotic bin packing.
In contrast, robotic bin packing requires modeling object geometry and determining feasible object poses to improve spatial utilization within a container.

For decades, robotic bin packing has been treated as a geometric optimization problem with a goal of maximizing packing compactness based on object shapes, poses, and container states~\cite{karabulut2004hybrid, kang2012hybrid,crainic2008extreme,liu2015hape3d, lamas2022voxel, jiang2012learning, ramos2016container, wang2019stable, wang2021dense, pan2023sdf, hu2020tap, zhao2021online, verma2020generalized, zhang2021attend2pack, yang2023heuristics, pan2023adjustable, huang2022planning, zhao2023learning, yang2021packerbot}.
Existing methods can be grouped into three categories: optimization-based approaches, packing heuristics, and reinforcement-learning-based methods.
However, most of these approaches focus on geometric modeling, typically neglecting object properties
such as fragility, edibility, and chemical compatibility.

\vspace{+5pt}
\noindent\textbf{Optimization-Based Methods.} \ 
Several optimization-based approaches have been proposed for robotic bin packing, including genetic algorithms~\cite{ramos2016container, karabulut2004hybrid, kang2012hybrid}, tabu search~\cite{crainic2008extreme}, simulated annealing~\cite{liu2015hape3d}, and integer linear programming~\cite{lamas2022voxel, jiang2012learning}.
Given a set of objects and a container, these methods evaluate object ordering and placement candidates offline to optimize packing compactness.
However, their use in a practical system is limited by
(i) the computational cost of planning an entire packing sequence offline and (ii) the need for replanning when executed object poses deviate from planned poses.
Although some earlier works study container balance by distributing heavy objects across horizontal positions~\cite{moon2014container,de2013two}, the aspects of object fragility and pairwise incompatibility remain underexplored.

\begin{figure*}
    \centering
    \includegraphics[width=0.99\linewidth, height=0.33\linewidth]{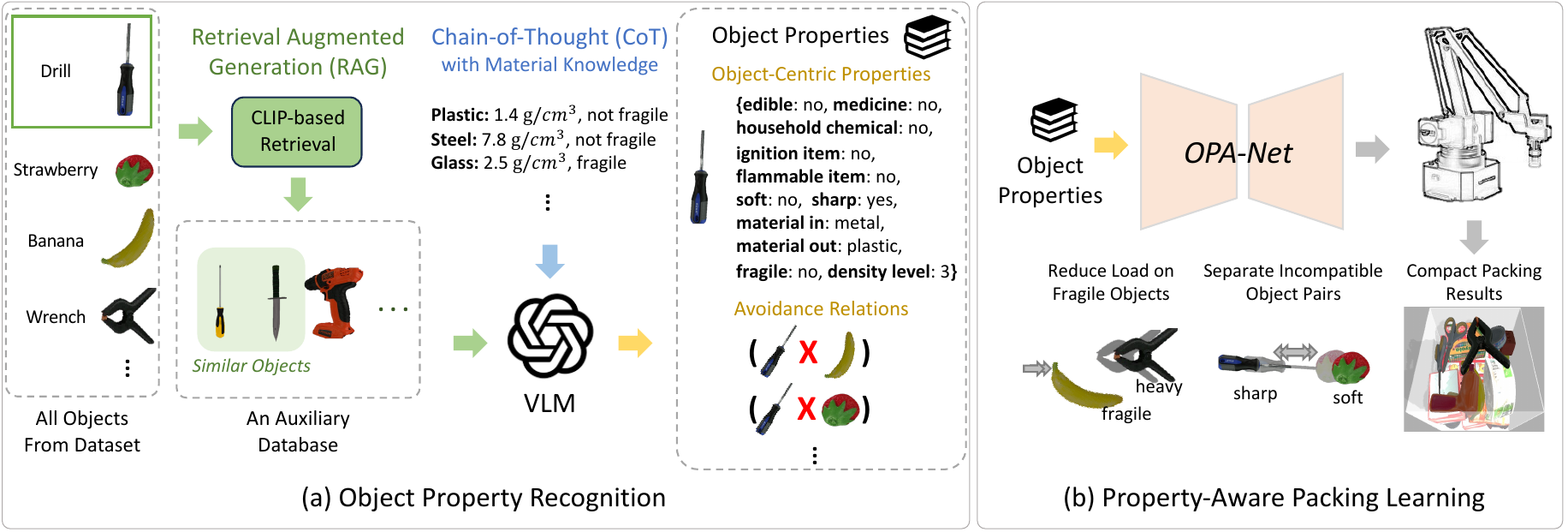}
    \vspace{-8pt}
     \caption{Illustration of OPA-Pack. 
     The framework consists of two stages: (a) object property recognition and (b) property-aware packing learning.
     In the first stage, a Vision-Language Model (VLM) is employed to predict the object-centric properties and the avoidance relations between object pairs.
     Specifically, for each query object, CLIP-based Retrieval-Augmented Generation (RAG) is designed to retrieve similar annotated objects from an auxiliary database as references.
     In addition, Chain-of-Thought (CoT) reasoning is designed to incorporate prior knowledge of typical material categories to guide the prediction of object-centric properties.
     Furthermore, we deduce avoidance relations between object pairs to determine incompatible object pairs that should be separated during packing.
     In the second stage, OPA-Net is formulated for property-aware packing learning to reduce the load on fragile objects and separate incompatible object pairs while preserving packing compactness.}
     \label{fig:framework}
\end{figure*}

\vspace{+5pt}
\noindent\textbf{Packing Heuristics.} \
To support online planning for robotic bin packing, several heuristics have been developed~\cite{karabulut2004hybrid, ramos2016container, wang2019stable, wang2021dense, pan2023sdf}.
These heuristics use specialized objective functions to select objects and determine feasible placement configurations online.
For instance, the deepest bottom-left heuristic~\cite{karabulut2004hybrid} prioritizes the deepest, bottom-most, left-most position; the empty-maximal-space heuristic~\cite{ramos2016container} prioritizes placements that maintain a maximal cuboid-shaped empty space; the heightmap-minimization method~\cite{wang2019stable, wang2021dense} prioritizes placements with minimal heightmap increments, thereby reducing unused space beneath placed objects.
The SDF-Minimization method~\cite{pan2023sdf} selects objects and placements by minimizing the signed distance field values to improve packing compactness.
These heuristics are computationally efficient for real-world robotic bin packing.
However, they depend on carefully crafted objectives, and their greedy nature limits long-term optimization of packing performance.
Moreover, existing packing heuristics generally neglect object-property constraints in the planning procedure.

\vspace{+5pt}
\noindent\textbf{Reinforcement Learning (RL).} \ 
RL methods support long-term optimization of packing performance over multiple steps.
Multiple objectives can be jointly incorporated in a unified reward function.
Recent studies have applied RL to robotic bin packing~\cite{hu2020tap,zhao2021online,verma2020generalized,zhang2021attend2pack,yang2023heuristics,pan2023adjustable, huang2022planning, zhao2023learning, yang2021packerbot}.
For instance, Verma et al.~\cite{verma2020generalized} proposed a Deep Q-Network (DQN) directly using packing compactness as the reward function.
Subsequent studies have investigated modified network architectures, including hierarchical DQN~\cite{huang2022planning} and dueling DQN~\cite{zhao2023learning}.
Other studies incorporate stability-related rewards~\cite{hu2020tap, zhao2021online, huang2022planning} or packing-heuristic-guided rewards~\cite{yang2023heuristics}.
However, existing RL-based packing algorithms primarily focus on geometric aspects, and generally neglect object properties in their modeling.

\vspace{+5pt}

Overall, prior robotic bin packing studies primarily address geometric aspects of the task, while object properties relevant to packing safety and compatibility have received limited attention.
As a result, existing methods may produce configurations in which heavy objects are placed above fragile items or incompatible objects are placed in adjacent positions, increasing the risk of product damage or contamination during packing and storage.
To address this limitation, this work introduces an object-property-aware formulation for robotic bin packing that explicitly encourages spatial separation between incompatible object pairs and reduces the load applied to fragile objects while preserving packing compactness.
This formulation extends robotic bin packing beyond geometric compactness by incorporating property-related constraints relevant to everyday object packing.

\section{Overview of OPA-Pack}
OPA-Pack is formulated as a two-stage framework comprising (a) object property recognition and (b) property-aware packing learning.
Figure~\ref{fig:framework} provides an illustration of the proposed framework.

First, in object property recognition, we predict object-centric properties and identify avoidance relations for each object pair. 
The resulting annotations constitute the OPA dataset.
An auxiliary database of human-annotated object-centric properties is used as the reference set.
For each query object, a Vision-Language Model (VLM) is employed to infer its object-centric properties. 
A retrieval-augmented generation scheme constructs in-context examples based on similar objects retrieved from the auxiliary database.
In addition, chain-of-thought reasoning is used to incorporate prior knowledge of typical material properties.
These components are designed to improve object-centric property prediction by providing object-level references and prior knowledge of material properties.
The predicted object-centric properties are utilized in a rule-based procedure to identify avoidance relations for all object pairs in the dataset.

Second, in property-aware packing learning, the Object-Property-Aware Network, termed OPA-Net, is constructed and trained using deep Q-learning to learn a property-aware packing strategy.
The packing process is formulated as a sequential decision-making problem.
At each step, OPA-Net selects a candidate object from the buffer and determines its orientation and placement in the container.
The model is trained with an OPA reward function comprising three reward terms that simultaneously reduce load on fragile objects, separate incompatible pairs, and optimize packing compactness.

\begin{figure*}
    \centering
    \includegraphics[width=\linewidth, height=0.58\linewidth]{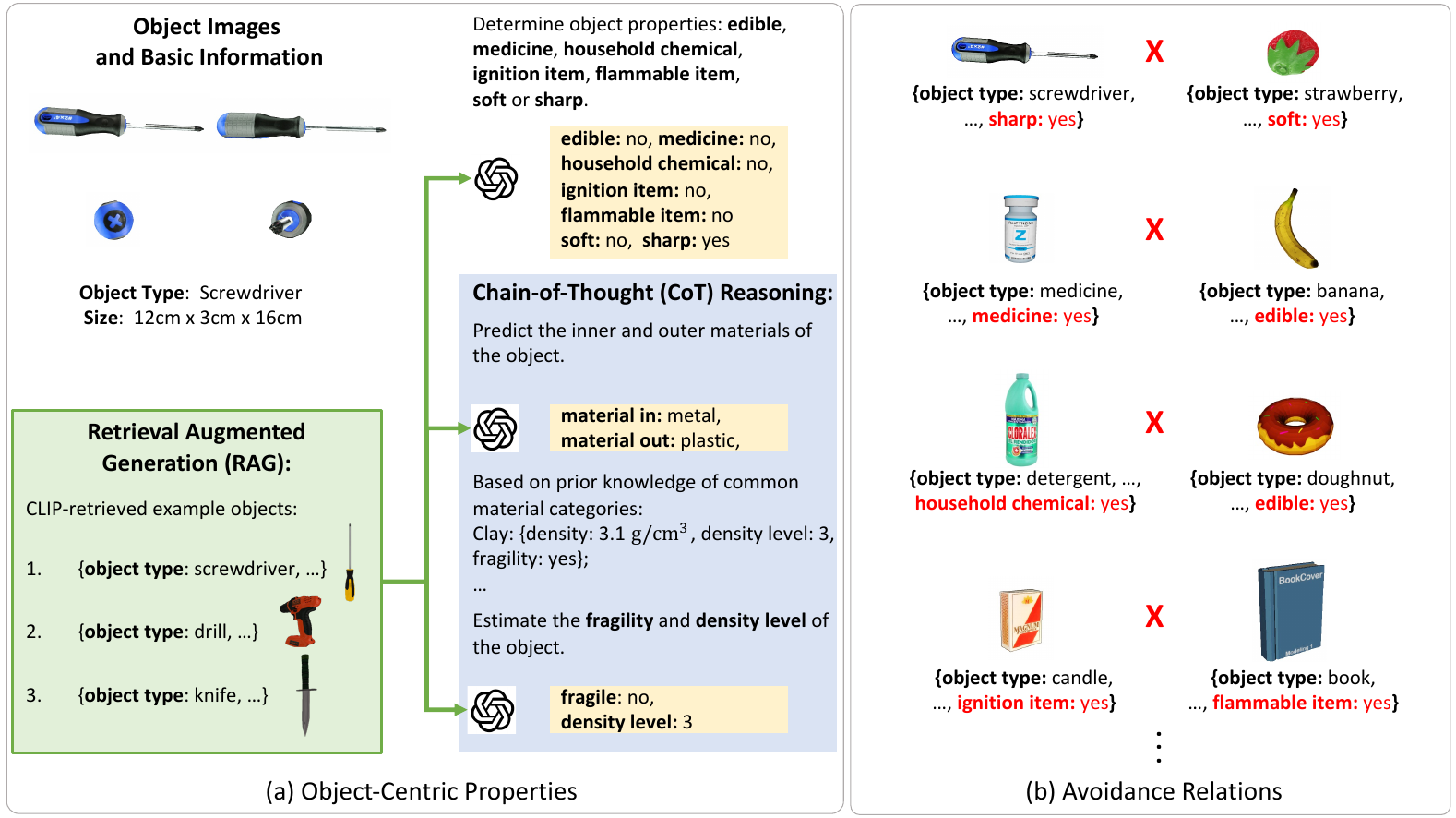}
     \caption{The object property recognition scheme predicts (a) object-centric properties and (b) avoidance relations. 
     For each query object, four orthogonal views of the object, its class name, and its size are used as input to predict eleven object-centric properties (yellow boxes).
     To enhance prediction performance, we perform RAG by retrieving example objects from an auxiliary database based on the similarity in CLIP features (green box).
     Five semantic properties and two physical properties of the query object are predicted.
     To predict the fragility and density level of the query object, we perform CoT reasoning by first predicting the inner and outer materials of the query object, and then leveraging the prior knowledge of typical material properties (Table~\ref{tab:density}) to guide the prediction (blue box).
     In addition, the avoidance relations between object pairs are predicted, including (i) sharp objects vs. soft objects, (ii) medicines vs. edible objects, (iii) household chemicals vs. edible objects, and (iv) ignition items vs. flammable items.
     }
     \label{fig:llm}
\end{figure*}

\section{Object Property Recognition}
\label{sec:dataset}

This section introduces the object property recognition procedure, which predicts object-centric properties for each object and derives avoidance relations for each object pair.
Figure~\ref{fig:llm} shows an illustration of the object property recognition scheme.

We obtain object-centric properties for each object based on a Vision-Language Model (VLM).
Retrieval-augmented generation is designed to construct in-context examples based on similar objects retrieved from an auxiliary database.
Chain-of-thought reasoning is designed to incorporate prior knowledge of typical material properties.
A rule-based procedure then derives avoidance relations from the predicted object-centric properties, indicating whether each object pair should be spatially separated during packing.
The resulting OPA dataset contains 1,032 objects annotated with object-centric properties and pairwise avoidance relations.

\vspace{+5pt}
\noindent\textbf{Object-Centric Property Prediction.} \
For each object, eleven object-centric properties are predicted, including five semantic properties (\ie, medicines, edibility, household chemicals, ignition items, and flammable items) and six physical properties (\ie, softness, sharpness, inner material, outer material, fragility, and density level).
Figure~\ref{fig:llm}(a) illustrates the object-centric property prediction process.
These properties are not always directly observable from visual and geometric information, requiring the integration of available object information with prior knowledge for property inference.
GPT-4o~\cite{achiam2023gpt} is used as the VLM because it supports multimodal inputs, enabling property prediction from rendered object views and associated textual information.
The VLM input includes four orthogonal views rendered from the 3D model of each object, the object category and the object size information.
We perform Retrieval-Augmented Generation (RAG)~\cite{zhao2024retrieval,lewis2020retrieval} to provide in-context object examples, and Chain-of-Thought (CoT) reasoning~\cite{wei2022chain,feng2024towards,chu2023survey} to incorporate prior knowledge of material properties into the prediction process.

\begin{table}[h]
    \centering
    \caption{Common material categories used for estimating density levels and fragility.
    }
    \resizebox{\linewidth}{!}{
    \begin{tabular}{c|c|c|c}
        \hline
        \textbf{Material $k$} & \textbf{Density Level $d_{k}$} & \textbf{Density $\sigma_{k}$ ($\mathrm{g}/\mathrm{cm}^{3}$}) & \textbf{Fragility} \\
        \hline
        \textbf{Air}    & 0   & 0.0  &  No\\
        \hline
        \textbf{Bread}    & 1   & 0.3  &  Yes  \\
        \hline
         \textbf{Biscuit} & 1 & 0.4  & Yes \\
        \hline
         \textbf{Wood}    & 1   & 0.4   &  No \\
        \hline
         \textbf{Cardboard} & 1 & 0.6  & No\\
        \hline
         \textbf{Wax}    & 2   &  0.9 & Yes\\
        \hline
         \textbf{Water}    & 2   & 1.0   & No \\
        \hline
         \textbf{Fruits}   & 2   &  1.0 & Yes\\
        \hline
         \textbf{Synthetic (Nylon)} & 2 & 1.1 &  No\\
        \hline
        \textbf{Paper}   & 2   &  1.2  & No\\
        \hline
         \textbf{Plastic}    & 2  & 1.4 & No\\
        \hline
        \textbf{Glass}    & 3   & 2.5   &  Yes\\
        \hline
         \textbf{Clay}   & 3   &  3.1  & Yes \\
        \hline
        \textbf{Ceramic}    & 4   & 4.2  &  Yes \\
        \hline
        \textbf{Steel}    & 5   & 7.8   &  No\\
        \hline
    \end{tabular}}
    \label{tab:density}
\end{table}

To facilitate RAG, an auxiliary database comprising 40 example objects from the YCB dataset~\cite{ycb} is constructed as the retrieval source.
Each database entry includes the 3D model of an object and manual annotations of the object's class name, size, and object-centric properties.
For the query object, we retrieve the top-10 most similar example objects from this database using the CLIP features~\cite{radford2021learning}.
Specifically, we extract the CLIP features using the front views of the query object and all example objects.
Next, we identify the example objects with CLIP features that have the largest cosine similarity to those of the query object.
The annotations of the retrieved example objects are included as in-context examples for predicting the five semantic properties and two physical properties, \ie, softness and sharpness.

Fragility and density level are less directly determined by visual appearance and retrieved examples, as both attributes depend on the material composition of the query object.
A two-stage CoT reasoning strategy is therefore used for predicting these two properties.
In the first stage, the inner and outer materials of the query object are predicted.
In the second stage, prior knowledge of material properties and the definitions of density levels are further utilized to infer the fragility and density level of the object.
Table~\ref{tab:density} summarizes prior knowledge about the properties of predefined materials.
For materials not listed in the predefined table, the VLM predicts fragility and density levels based on prior knowledge.

\vspace{+5pt}
\noindent\textbf{Avoidance Relation Prediction.} \
Pairwise avoidance relations are derived from the predicted object-centric properties using a rule-based procedure.
Each avoidance relation specifies whether an object pair should be separated during packing.
Four types of avoidance relations are considered: 
(i) sharp objects and soft objects, because sharp objects may puncture the soft objects; 
(ii) medicines and edible objects; (iii) household chemicals and edible objects, because these two combinations may create contamination risks; and (iv) ignition items and flammable items, because close placement may increase fire risk.
Figure~\ref{fig:llm}(b) illustrates the prediction of avoidance relations between pairs of objects.
An incompatible pair is defined as two objects with an avoidance relation.

\begin{figure}[]
    \centering
    \includegraphics[width=\linewidth, height=0.38\linewidth]{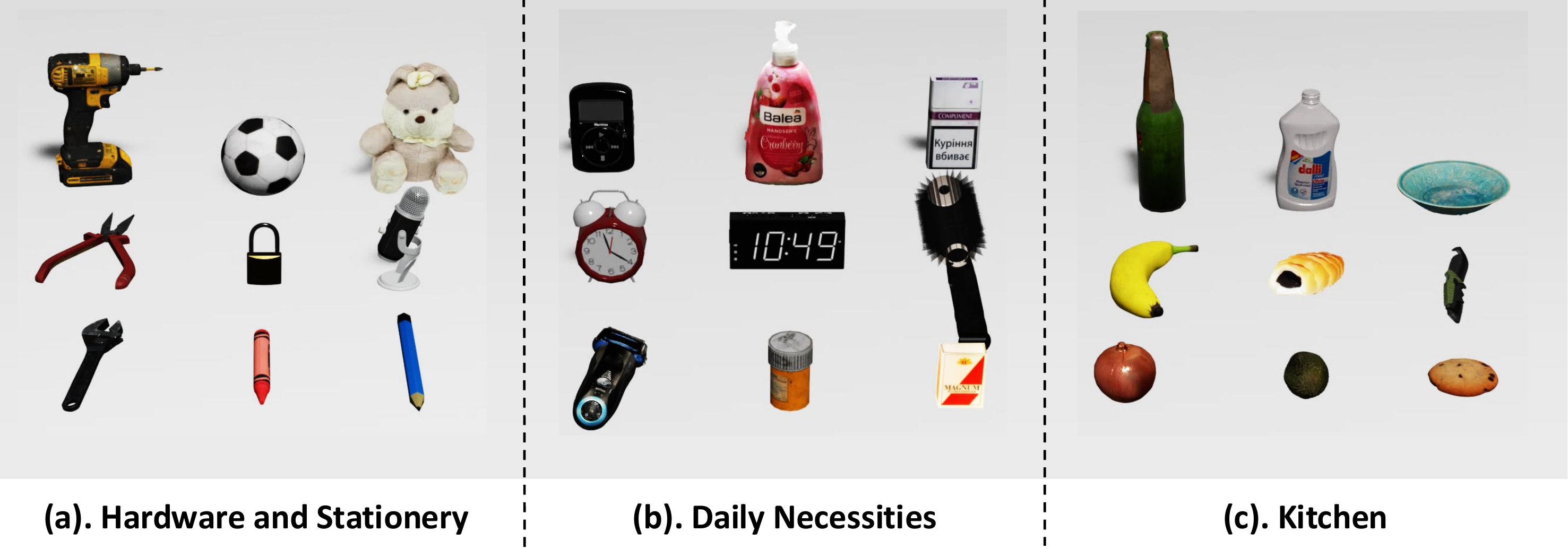}
     \caption{Example objects from the OPA dataset divided into three subsets.}
     \label{fig:object}
\end{figure}

\vspace{+5pt}
\noindent\textbf{OPA Dataset Construction.} \
We collect everyday objects from the Objaverse-v1 dataset~\cite{deitke2023objaverse} using object class names to construct the OPA dataset.
Based on manual inspection, objects are retained if they correspond to everyday items and have a longest bounding-box dimension below \(30\,\mathrm{cm}\), resulting in 1,032 objects.
The objects are partitioned into three subsets based on their usage contexts: \textit{Kitchen}, \textit{Hardware and Stationery}, and \textit{Daily Necessities}.
Figure~\ref{fig:object} shows representative objects from the dataset.
For each object, the 3D geometry and object class name are retained from the Objaverse-v1 dataset.
The eleven object-centric properties are predicted using the proposed property recognition pipeline, and pairwise avoidance relations are derived from these properties using the four avoidance rules defined above.
The dataset will be released upon publication.

\section{Property-aware Packing Learning}
\label{sec:rl}

OPA-Net uses the estimated object properties for property-aware packing learning.
The network is based on deep Q-learning~\cite{hu2020tap,huang2022planning,zhao2023learning}, which supports long-horizon decision making and joint optimization of multiple packing objectives.

\subsection{Deep Q-learning Formulation for Packing}
The packing process sequentially selects an object from a buffer $\mathcal{B}$ and determines its placement in the container.
This sequential decision-making problem is formulated as a Markov Decision Process with packing step $t \in \mathbb{N}_{+}$, state $\mathbf{s}_{t}$, action $\mathbf{a}_{t}$, and reward function $R(\mathbf{s}_{t}, \mathbf{a}_{t})$.
The state $\mathbf{s}_{t}$ includes the geometry and pose of each packed object and the geometry of the buffered objects.
The action $\mathbf{a}_{t}$ specifies the selected object and its placement at packing step $t$.
The reward function $R(\mathbf{s}_{t}, \mathbf{a}_{t})$ evaluates the packing quality resulting from executing action $\mathbf{a}_{t}$ in state $\mathbf{s}_{t}$.
A higher reward value indicates better packing quality.

Packing planning aims to maximize the expected cumulative reward over the packing horizon rather than the immediate reward at each step.
A Q-function~\cite{zhao2021learning, huang2022planning, watkins1992q} $Q(\mathbf{s}_{t}, \mathbf{a}_{t})$ is designed to estimate the expected cumulative return of taking action $\mathbf{a}_{t}$ in state $\mathbf{s}_{t}$ and following the policy thereafter.
The Q-value is updated using the standard Q-learning update rule:
\begin{equation}
\label{eq:q_val}
\begin{split}
     Q_{\mathrm{new}}(\mathbf{s}_{t}, \mathbf{a}_{t}) = & (1 - \alpha) \cdot Q(\mathbf{s}_{t}, \mathbf{a}_{t}) \\
     & + \alpha \cdot ( R(\mathbf{s}_{t}, \mathbf{a}_{t}) + \gamma \cdot \max_{\mathbf{a}' \in \mathcal{A}} Q(\mathbf{s}_{t+1}, \mathbf{a}')),
\end{split}
\end{equation}
where $\gamma$ is the discount factor, $\alpha$ is the constant learning rate, $\mathcal{A}$ is the set of all candidate packing actions, and $Q_{\mathrm{new}}(\mathbf{s}_{t}, \mathbf{a}_{t})$ denotes the updated Q-value.

Conventional Q-learning methods store the Q-values in a table~\cite{watkins1992q} and update them using Eq.~\eqref{eq:q_val}.
However, tabular Q-learning becomes infeasible for high-dimensional state and action spaces.
Therefore, a Deep Q-Network (DQN)~\cite{zhao2021learning, huang2022planning} is trained to estimate $Q(\mathbf{s}_{t}, \mathbf{a}_{t})$ from the scene observation $\operatorname{obs}(\mathbf{s}_{t})$.
The predicted Q-value is written as $\widetilde{Q}(\operatorname{obs}(\mathbf{s}_{t}), \mathbf{a}_{t})$.
$\operatorname{obs}(\mathbf{s}_{t})$ is used instead of $\mathbf{s}_{t}$, because the packing scene is partially observable; the full states of occluded objects may be unavailable.
The DQN is trained by minimizing the squared temporal-difference (TD) error between the target and the predicted Q-values:
\begin{equation}
\label{eq:loss}
    \mathcal{L} =  (Q_{\mathrm{tar}}(\mathbf{s}_{t}, \mathbf{a}_{t}) - \widetilde{Q}(\operatorname{obs}(\mathbf{s}_{t}), \mathbf{a}_{t}))^2,
\end{equation}
where $Q_{\mathrm{tar}}(\mathbf{s}_{t}, \mathbf{a}_{t})$ denotes the target Q-value obtained following standard DQN~\cite{huang2022planning, zhao2023learning}. 
At test time, the action is selected according to $\mathbf{a}_{t}^* = \arg\max_{\mathbf{a}' \in \mathcal{A}} (\widetilde{Q}(\operatorname{obs}(\mathbf{s}_{t}), \mathbf{a}'))$.

The standard Dueling Deep Q-Network (Dueling DQN)~ \cite{wang2016dueling} is employed to train OPA-Net.
Dueling DQN separates $\widetilde{Q}(\operatorname{obs}(\mathbf{s}_{t}), \mathbf{a}_{t})$ into a state-value function $\widetilde{V}(\operatorname{obs}(\mathbf{s}_{t}))$ that estimates the value of the current scene observation
and an advantage function $\widetilde{A}(\operatorname{obs}(\mathbf{s}_{t}), \mathbf{a}_{t})$ that measures the relative benefit of action $\mathbf{a}_{t}$ at packing step $t$.
This relationship is expressed as 
\begin{equation}
\begin{split}
\label{eq:duel_q}
\widetilde{Q}(\operatorname{obs}(\mathbf{s}_{t}), \mathbf{a}_{t}) = & \widetilde{V}(\operatorname{obs}(\mathbf{s}_{t})) + \widetilde{A}(\operatorname{obs}(\mathbf{s}_{t}), \mathbf{a}_{t}) \\ & - {\operatorname{mean}}_{\mathbf{a}' \in \mathcal{A}}(\widetilde{A}(\operatorname{obs}(\mathbf{s}_{t}), \mathbf{a}')),
\end{split}
\end{equation}
where ${\operatorname{mean}}_{\mathbf{a}' \in \mathcal{A}} (\cdot)$ denotes the mean pooling operation across all candidate actions, and $\mathcal{A}$ is the set of all candidate actions.

Existing deep Q-learning methods for robotic bin packing primarily optimize packing compactness.
OPA-Net incorporates object-property-aware planning into packing learning through two components: (i) property-aware observations $\operatorname{obs}(\mathbf{s}_{t})$ that capture not only the geometries but also the properties of both packed and buffered objects, and (ii) a reward function $R(\mathbf{s}_{t}, \mathbf{a}_{t})$ that optimizes packing compactness while reducing load on fragile objects and separating incompatible object pairs.

\subsection{OPA-Net}
\label{sec:opa_net}
\begin{figure}[]
    \centering
    \includegraphics[width=\linewidth, height=0.95\linewidth]{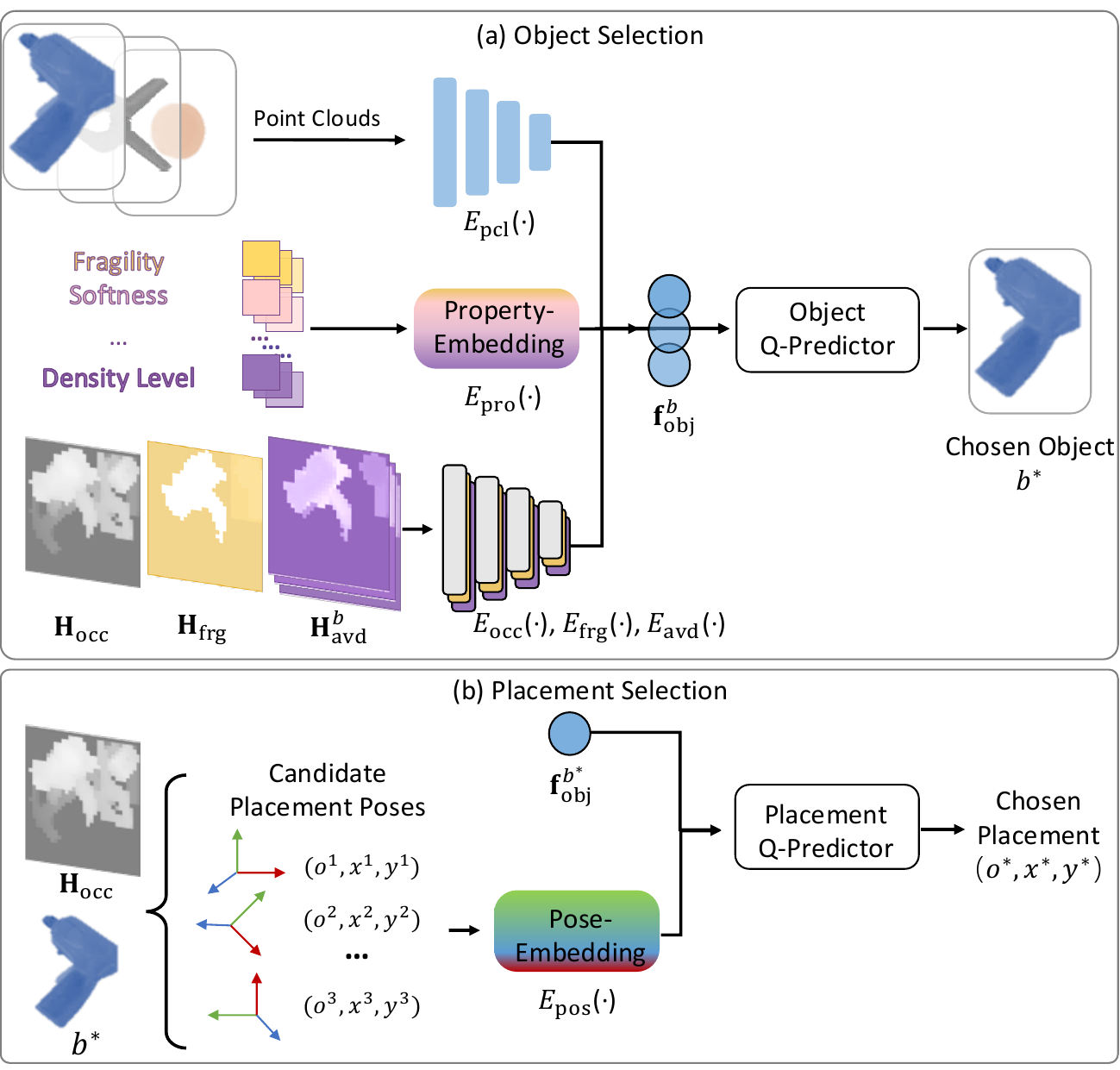}
     \caption{Illustration of OPA-Net performing (a) object selection and (b) placement selection. 
     For (a), OPA-Net uses a point cloud encoder $E_{\mathrm{pcl}}$ and a property-embedding layer $E_{\mathrm{pro}}$ to encode geometric and property information for each object $b$ in the buffer $\mathcal{B}$.
     It constructs occupancy, fragility, and avoidance heightmaps $(\mathbf{H}_{\mathrm{occ}}, \mathbf{H}_{\mathrm{frg}}, \mathbf{H}_{\mathrm{avd}}^b)$ to represent the packing conditions of the container and then learns heightmap features through the occupancy, fragility and avoidance heightmap encoders $E_{\mathrm{occ}}$, $E_{\mathrm{frg}}$, and $E_{\mathrm{avd}}$ to form the object feature $\mathbf{f}_{\mathrm{obj}}^{b}$.
     OPA-Net uses an object Q-predictor to select the object $b^*$ for packing.
     For (b), OPA-Net extracts candidate poses for $b^*$ from $\mathbf{H}_{\mathrm{occ}}$ and employs a pose-embedding layer to encode the candidate poses.
     A placement Q-predictor selects the placement with the highest predicted Q-value based on the object features $\mathbf{f}_{\mathrm{obj}}^{b^*}$ and the features of the candidate poses.
     }
     \label{fig:network_archi}
\end{figure}

OPA-Net integrates object-property encoding and property-aware container heightmaps for object-property-aware packing learning. 
The network determines the packing action $\mathbf{a}_{t} = (b^*, o^*, x^*, y^*)$ at each packing step, where $b^*$ is the selected object from the buffer $\mathcal{B}$, and $o^*, x^*, y^*$ are the selected orientation and horizontal placement coordinates for the object, respectively.
The vertical location $z^*$ is determined by the support height at the selected horizontal placement under gravity.
Three encoders process the geometry and properties of each candidate object $b \in \mathcal{B}$ together with the container heightmaps, producing the object feature $\mathbf{f}_{\mathrm{obj}}^{b}$.
An object Q-predictor estimates the Q-value of each candidate object $b \in \mathcal{B}$, from which $b^*$ is selected by maximizing the predicted Q-value.
Given $b^*$ and the container heightmap, a set of candidate placements $(o, x, y)$ is generated, and the corresponding pose features are extracted.
A placement Q-predictor estimates the Q-values of the candidate placements from the object feature $\mathbf{f}_{\mathrm{obj}}^{b^*}$ and the pose features, and $(o^*, x^*, y^*)$ is selected as the placement with the maximum predicted Q-value.
Figure~\ref{fig:network_archi} illustrates the architecture of OPA-Net.

\vspace{+5pt}

\noindent\textbf{Object Geometry Encoding.}\
For each candidate object $b$ in the buffer $\mathcal{B}$, we sample a point cloud $\mathbf{P}^b$ that captures the 3D geometry of the object.
PointNet~\cite{qi2017pointnet}, denoted by $E_{\mathrm{pcl}}(\cdot)$, is used as the point-cloud feature extractor:
\begin{equation}
    \mathbf{f}_{\mathrm{pcl}}^{b} = E_{\mathrm{pcl}}(\mathbf{P}^b),
\end{equation}
where $\mathbf{f}_{\mathrm{pcl}}^{b}$ denotes the point-cloud feature of the object.

\vspace{+5pt}

\noindent\textbf{Object Property Encoding.} \
Fragility, softness, and sharpness are represented as binary scalars, whereas estimated density and volume are represented as real-valued scalars.
For each object with density level $d$, the estimated density $\hat{\sigma}_d$ is defined as the mean density of all materials $k$ in level $d$, \ie, $ \hat{\sigma}_d = \operatorname{mean}_{\{k:\, d_k = d\}}(\sigma_k)$ (Table~\ref{tab:density}).
The input object properties for object $b$ are concatenated to form a property vector $\mathbf{x}_{\mathrm{pro}}^{b}$ and are encoded by
\begin{equation}
    \mathbf{f}_{\mathrm{pro}}^{b} = E_{\mathrm{pro}}(\mathbf{x}_{\mathrm{pro}}^{b}),
\end{equation}
where $E_{\mathrm{pro}}(\cdot)$ denotes the property-embedding layer, which is a multi-layer perceptron, and $\mathbf{f}_{\mathrm{pro}}^{b}$ denotes the property feature.

\vspace{+5pt}

\noindent\textbf{Container Heightmap Encoding.} \
OPA-Net takes three container heightmaps as input: an occupancy heightmap $\mathbf{H}_{\mathrm{occ}}$, a fragility heightmap $\mathbf{H}_{\mathrm{frg}}$, and an avoidance heightmap $\mathbf{H}_{\mathrm{avd}}^b$ for each candidate object $b$, as illustrated in the bottom-left of Figure~\ref{fig:network_archi}.

The occupancy heightmap $\mathbf{H}_{\mathrm{occ}}$ is generated from a depth camera mounted above the container to record the occupied height at each grid location.
The fragility heightmap $\mathbf{H}_{\mathrm{frg}}$ and the avoidance heightmap $\mathbf{H}_{\mathrm{avd}}^b$ are constructed from the locations of previously packed fragile objects and previously packed objects that have avoidance relations with candidate object $b$, respectively.
The object properties $\mathbf{x}_{\mathrm{pro}}^{i}$, packing orientation $o_i$, and packing location $(x_i, y_i, z_i)$ are recorded for each packed object $i$ to obtain the fragility heightmap $\mathbf{H}_{\mathrm{frg}}$ and the avoidance heightmap $\mathbf{H}_{\mathrm{avd}}^b$.

The fragility heightmap encodes the top heights of previously packed fragile objects. 
It is initialized as an all-zero 2D matrix with the same dimensions as the occupancy heightmap $\mathbf{H}_{\mathrm{occ}} \in \mathbb{R}^{W \times L}$, which is written as
\begin{equation}
    \mathbf{H}_{\mathrm{frg}}^{(0)} = \mathbf{0}_{W \times L},
\end{equation}
where $\mathbf{H}_{\mathrm{frg}}^{(0)}$ denotes the initialized fragility heightmap; $W$ and $L$ denote the width and length of the container in grid cells, respectively.
The fragility heightmap is updated iteratively to estimate the top surface formed by all packed fragile objects:

\begin{equation}
\begin{aligned}
\operatorname{crop}(\mathbf{H}_{\mathrm{frg}}^{(i)},\Omega_i)
= &\operatorname{max_{e}}
(\operatorname{crop}(\mathbf{H}_{\mathrm{frg}}^{(i-1)}, \Omega_i), \mathbf{H}^{o_i}_{i} + \mathbf{Z}_{i}), \\
\Omega_i = &\{(x',y') \, | \, \tilde x_i \le x' < \tilde x_i+w_i^{o_i},\\
& \hspace{+40pt} \tilde y_i \le y' < \tilde y_i+l_i^{o_i}, x', y' \in \mathbb{N}^{+}\}, \\
i = &1, 2, ..., |\mathcal{F}|, 
\end{aligned}
\end{equation}
where $i$ indexes a packed fragile object; $\Omega_i$ represents the rectangular index region covered by object $i$ on the horizontal plane, determined by the grid indices $(\tilde x_i,\tilde y_i)$ corresponding to its horizontal placement coordinates $(x_i,y_i)$, the selected orientation $o_i$, and its width $w_i^{o_i}$ and length $l_i^{o_i}$ in grid cells;
$\operatorname{crop}(\cdot,\Omega_i)$ denotes the cropping operation that extracts the sub-heightmap from an input heightmap indexed by $\Omega_i$;
$\mathbf{H}_{\mathrm{frg}}^{(i)}$ retains the values of  $\mathbf{H}_{\mathrm{frg}}^{(i-1)}$ outside $\Omega_i$.
$\mathbf{H}^{o_i}_{i}\in(\mathbb R\cup \{-\infty\})^{w_i^{o_i}\times l_i^{o_i}}$ denotes the top-down heightmap of object $i$ at orientation $o_i$, with $-\infty$ assigned to pixels not occupied by the object;
suppose $z_i$ is the vertical placement coordinate of object $i$,
$\mathbf{Z}_{i} \in \mathbb{R}^{w_i^{o_i} \times l_i^{o_i}}$ denotes a constant sub-heightmap whose entries are all equal to $z_i$;
therefore, $\mathbf{H}^{o_i}_{i} + \mathbf{Z}_{i}$ denotes the estimated top surface of the packed fragile object $i$ at the selected placement;
$\operatorname{max_{e}}(\cdot, \cdot)$ denotes the element-wise max operation;
$\mathcal{F}$ denotes the set of all packed fragile objects;
and the generated fragility map is given by $\mathbf{H}_{\mathrm{frg}} = \mathbf{H}_{\mathrm{frg}}^{(|\mathcal{F}|)}$.

The avoidance heightmap $\mathbf{H}_{\mathrm{avd}}^b$ is constructed analogously to estimate the top surface formed by the packed objects with avoidance relations with the candidate object $b$.
Unlike $\mathbf{H}_{\mathrm{occ}}$ and $\mathbf{H}_{\mathrm{frg}}$, different candidate objects $b$ may have distinct $\mathbf{H}_{\mathrm{avd}}^b$, because avoidance relations with previously packed objects depend on $b$.

The three heightmaps are each encoded by a convolutional neural network (CNN), written as follows:
\begin{equation}
\begin{split}
    \mathbf{f}_{\mathrm{occ}} & =  E_{\mathrm{occ}}( \mathbf{H}_{\mathrm{occ}}), \\ \mathbf{f}_{\mathrm{frg}} & = E_{\mathrm{frg}}(\mathbf{H}_{\mathrm{frg}}), \\  \mathbf{f}_{\mathrm{avd}}^b & = E_{\mathrm{avd}}(\mathbf{H}_{\mathrm{avd}}^b),
\end{split}
\end{equation}
where $E_{\mathrm{occ}}(\cdot)$, $E_{\mathrm{frg}}(\cdot)$, and $E_{\mathrm{avd}}(\cdot)$ denote the CNN encoders for the occupancy, fragility, and avoidance heightmaps, respectively, and $\mathbf{f}_{\mathrm{occ}}$, $\mathbf{f}_{\mathrm{frg}}$, and $\mathbf{f}_{\mathrm{avd}}^b$ denote the resulting feature maps.

\vspace{+5pt}

\noindent\textbf{Object Q-Predictor.} \
For each candidate object $b$, the object geometric feature $\mathbf{f}_{\mathrm{pcl}}^{b}$, object property feature $\mathbf{f}_{\mathrm{pro}}^{b}$, and container features $\mathbf{f}_{\mathrm{occ}}$, $\mathbf{f}_{\mathrm{frg}}$, and $\mathbf{f}_{\mathrm{avd}}^b$ are used to predict the object-selection Q-value.
We adopt the dueling DQN architecture where the Q-value is obtained from the state value $\widetilde{V}_{\mathrm{obj}}$ and the advantage value $\widetilde{A}_{\mathrm{obj}}^b$, which is written as
\begin{equation}
\begin{split}
\mathbf{f}_{\mathrm{obj}}^b
&=
\operatorname{concat}
(
\mathbf{f}_{\mathrm{pcl}}^b,
\mathbf{f}_{\mathrm{pro}}^b,
\mathbf{f}_{\mathrm{occ}},
\mathbf{f}_{\mathrm{frg}},
\mathbf{f}_{\mathrm{avd}}^b
), \\
\widetilde{V}_{\mathrm{obj}}
&=
\Phi_{\mathrm{V}}(\operatorname{mean}_{b \in \mathcal{B}}({\mathbf{f}_{\mathrm{obj}}^b})), \\
\widetilde{A}_{\mathrm{obj}}^b
&=
\Phi_{\mathrm{A}}
\left(
\mathbf{f}_{\mathrm{obj}}^b
\right),
\end{split}
\end{equation}
where $b$ is a candidate object from $\mathcal{B}$; 
$\mathbf{f}_{\mathrm{obj}}^b$ denotes the feature for object selection; 
$\widetilde{V}_{\mathrm{obj}}$ is the state value,
and $\widetilde{A}_{\mathrm{obj}}^b$ is the advantage associated with selecting object $b$;
the functions $\Phi_{\mathrm{V}}(\cdot)$ and $\Phi_{\mathrm{A}}(\cdot)$ are multi-layer perceptrons (MLPs).
The Q-value for object selection is obtained following Eq.~\eqref{eq:duel_q}, and the selected object $b^*$ is determined as the candidate object with the highest Q-value.

\vspace{+5pt}

\noindent\textbf{Candidate Placement Encoding.} \
Given the selected object $b^*$, its candidate placement set $\mathcal{E}$ is constructed, where each candidate placement consists of an orientation $o$ and a horizontal location $(x,y)$.
The candidate orientations are generated from the planar-stable poses of $b^*$, defined as poses in which the object remains stable on a planar surface~\cite{zhao2021learning}.
These poses are used as candidates because objects commonly rest in planar-stable poses prior to grasping, such as on a table or conveyor.
All planar-stable poses are considered during training to improve robustness.
During real-world testing, only the initial pose observed in the buffer area is considered.
For each planar-stable pose, eight in-plane rotations about the $z$-axis are generated to form the candidate orientations $o$.

For each orientation $o$, feasible horizontal locations $(x,y)$ for placing the selected object $b^*$ inside the container are evaluated.
For a fixed $(o,x,y)$, the vertical coordinate $z$ is uniquely determined by the lowest collision-free position under gravity. 
Therefore, each candidate placement is represented by $(o,x,y)$ in $\mathcal{E}$, and $z$ is computed subsequently~\cite{wang2019stable}.
To reduce the placement search space, only locally optimal positions $(x,y)$ are retained, where $b^*$ is in contact with previously packed objects or container walls.
These locally optimal positions are identified by locating all ``convex vertices'' of the placement area~\cite{zhao2021learning}. 
At such positions, further motion toward nearby obstacles is infeasible.

Each candidate placement $(o, x, y) \in \mathcal{E}$ is encoded by a pose embedding layer $E_{\mathrm{pos}}(\cdot)$:
\begin{equation}
\mathbf{f}_{\mathrm{pos}}^{b^*}[o,x,y]
= E_{\mathrm{pos}}(o,x,y),
\end{equation}
where $\mathbf{f}_{\mathrm{pos}}^{b^*}[o,x,y]$ denotes the pose feature of object $b^*$ for candidate placement $(o,x,y)$, and $\mathcal{E}$ denotes the set of all candidate placements of object $b^*$.

\vspace{+5pt}

\noindent\textbf{Placement Q-Predictor.} \
Given the selected object $b^*$, a placement Q-predictor selects the placement $(o^*, x^*, y^*)$ with the maximum Q-value.
The placement Q-predictor takes the object feature and pose features of $b^*$ as inputs to predict Q-values for candidate placements.
Following the dueling DQN formulation in Eq.~\eqref{eq:duel_q}, the placement Q-value is decomposed into a state value $\widetilde{V}_{\mathrm{place}}$ and an advantage value $\widetilde{A}_{\mathrm{place}}[o,x,y]$ associated with placement $(o,x,y)$:
\begin{equation}
\begin{split}
\mathbf{f}_{\mathrm{place}[o,x,y]}
&=
\operatorname{concat}
(\mathbf{f}_{\mathrm{obj}}^{b^*}, \mathbf{f}_{\mathrm{pos}}^{b^*}[o,x,y]), \\
\widetilde{V}_{\mathrm{place}}
&=
\Psi_V
(\operatorname{mean}_{(o,x,y) \in \mathcal{E}}
(\mathbf{f}_{\mathrm{place}}[o,x,y])), \\
\widetilde{A}_{\mathrm{place}}[o,x,y]
&=
\Psi_A
\left(
\mathbf{f}_{\mathrm{place}}[o,x,y]
\right),
\end{split}
\end{equation}
where $\mathbf{f}_{\mathrm{place}}[o,x,y]$ denotes the placement feature of the selected object $b^*$ associated with candidate placement $(o,x,y)$;
$\operatorname{mean}_{(o, x, y) \in \mathcal{E}}(\cdot)$ denotes the mean pooling operation across all candidate placements;
the functions $\Psi_{\mathrm{V}}(\cdot)$ and $\Psi_{\mathrm{A}}(\cdot)$ are MLPs.
The selected placement $(o^*,x^*,y^*)$ is the candidate with the maximum placement Q-value.

The pose features of the object are provided only to the placement Q-predictor and are not used by the object Q-predictor.
Although available candidate placements can affect the object decision, the point cloud feature $\mathbf{f}_{\mathrm{pcl}}^{b}$ provides object-level geometric information for object selection without explicitly evaluating all candidate placements.
This design reduces the computation required to evaluate all candidate placements for every candidate object $b$.
The effect of placement feasibility on object selection is learned implicitly from rewards through reinforcement learning based on the object-level features.

\subsection{The OPA Reward and Network Training}
\label{sec:reward_design}

The reward function for training OPA-Net incorporates object properties and packing compactness.
Following prior packing methods~\cite{hu2020tap, huang2022planning, zhao2023learning}, packing compactness is included as a reward term to encourage space utilization.
In addition, we introduce penalties for placing heavy objects on fragile items and for violating predefined avoidance relations to form an Object-Property-Aware (OPA) reward function.

Following~\cite{zhao2023learning}, we define the compactness term $C$ as the ratio between the total volume of packed objects and the container volume:
\begin{equation}
\label{eq:compactness}
    C = \frac{\sum_{i}{V_i}}{H \cdot W \cdot L},
\end{equation}
where $i$ indexes each packed object; $V_i$ is the volume of object $i$, measured by the number of occupied grid cells; and $H$, $W$, and $L$ represent the height, width, and length of the container measured in grid cells, respectively.
The range of the compactness term is $C \in [0,1]$, with a larger value indicating higher space utilization.

In addition, we formulate the fragility term $R_{\mathrm{frg}}$ and the avoidance term $R_{\mathrm{avd}}$.
The fragility term $R_{\mathrm{frg}}$ penalizes placing the selected object $b^*$ above fragile objects:
\begin{equation}
\label{eq:fragile_reward}
R_{\mathrm{frg}} = N_{\mathrm{frg}} \cdot \omega_{b^*},
\end{equation}
where $N_{\mathrm{frg}}$ denotes the number of fragile objects underneath the selected object $b^*$.
This term is scaled by $\omega_{b^*}$, the weight of $b^*$, to assign larger penalties for heavier objects.
The weight $\omega_{b^*}$ is computed as the product of the object volume and the estimated density associated with its predicted density level.

The avoidance term $R_{\mathrm{avd}}$ penalizes placements in which the selected object $b^*$ is packed within a distance threshold $\tau$ of at least one incompatible object that has an avoidance relation with $b^*$:
\begin{equation}
\label{eq:avoidance_reward}
R_{\mathrm{avd}} =
\begin{cases}
1, & \text{if $\operatorname{dist}(b^*, j) < \tau$ for any incompatible object $j$}, \\
0, & \text{otherwise},
\end{cases}
\end{equation}
where $\tau$ denotes the predefined distance threshold, and $\operatorname{dist}(\cdot,\cdot)$ denotes the surface distance between a pair of objects in the container.
The values of $R_{\mathrm{frg}}$ and $R_{\mathrm{avd}}$ are computed from the object positions available in the virtual environment.
The total OPA reward is formulated as a weighted combination of compactness and the two penalty terms:
\begin{equation}
\label{eq:reward}
    R_{\mathrm{total}} = \theta \cdot  C  - \lambda \cdot R_{\mathrm{frg}} - \beta \cdot R_{\mathrm{avd}},
\end{equation}
where $\theta$, $\lambda$, and $\beta$ are positive hyperparameters.

OPA-Net is trained using the squared temporal-difference (TD) error defined in Eq.~\eqref{eq:loss}.
The object and placement Q-predictors are trained using target Q-values corresponding to object selection and placement selection, respectively.

\begin{algorithm}
\caption{A Packing Step during Testing~\label{alg:inference}}
\begin{algorithmic}[1]
\Function{PackingStep}{$\mathbf{H}_{\mathrm{occ}}$, $\mathcal{B}$, $\mathcal{H}$}
\If {{$\mathcal{B} = \emptyset$ or $\operatorname{IsFull}(\mathbf{H}_{\mathrm{occ}})$}}
    \State \Return $(), \mathcal{H}$
\EndIf
\State $\mathbf{H}_{\mathrm{frg}} \gets \operatorname{GetFragileMap}(\mathcal{H})$
\For{$b \in \mathcal{B}$}
    \State $\mathbf{H}_{\mathrm{avd}}^b \gets \operatorname{GetAvoidanceMap}(\mathcal{H}, b)$
    \State $\mathbf{P}^b, \mathbf{x}_{\mathrm{pro}}^{b} \gets \operatorname{GetObjectInfo}(b)$
    \State $\mathbf{f}_{\mathrm{obj}}^{b} \gets \operatorname{ObjectEncode}(\mathbf{P}^b, \mathbf{x}_{\mathrm{pro}}^{b}, \mathbf{H}_{\mathrm{occ}}, \mathbf{H}_{\mathrm{frg}}, \mathbf{H}_{\mathrm{avd}}^b)$
\EndFor

\State $b^* \gets \textproc{ChooseObject}(\mathcal{B}, \{\mathbf{f}_{\mathrm{obj}}^{b}\}_{b \in \mathcal{B}})$
\State $o^*, x^*, y^*, z^* \gets \textproc{ChoosePlacement}(b^*,  \mathbf{H}_{\mathrm{occ}}, \mathbf{f}_{\mathrm{obj}}^{b^*})$

\State $\mathcal{T}_{\mathrm{place}} \gets (b^*, o^*, x^*, y^*, z^*)$

\State $\mathcal{H} \gets [\mathcal{H}; \mathcal{T}_{\mathrm{place}}]$

\State \Return $\mathcal{T}_{\mathrm{place}}, \mathcal{H}$

\EndFunction

\Function{ChooseObject}{$\mathcal{B}, \{\mathbf{f}_{\mathrm{obj}}^{b}\}_{b \in \mathcal{B}}$}
\State $Q_{\mathrm{obj}}^* \gets -\infty$
\For{$b \in \mathcal{B}$} 
    \State $Q_{\mathrm{obj}}^b \gets \operatorname{ObjectQ}(\mathbf{f}_{\mathrm{obj}}^{b})$
    \If {${Q_{\mathrm{obj}}^b > Q_{\mathrm{obj}}^*}$}
        \State $Q_{\mathrm{obj}}^* \gets Q_{\mathrm{obj}}^b$
        \State $b^* \gets b$
    \EndIf
\EndFor
\State \Return $b^*$
\EndFunction

\Function{ChoosePlacement}{$b^*,  \mathbf{H}_{\mathrm{occ}}, \mathbf{f}_{\mathrm{obj}}^{b^*}$}
\State $Q_{\mathrm{place}}^* \gets -\infty$
\State $\mathcal{E} \gets \operatorname{GetAllPlacements}(b^*, \mathbf{H}_{\mathrm{occ}})$ 
\State $\mathbf{f}_{\mathrm{place}} \gets \operatorname{PlacementEncode}(\mathbf{f}_{\mathrm{obj}}^{b^*}, \mathcal{E})$
\ForAll{$(o,x,y) \in \mathcal{E}$}
    \State $Q_{\mathrm{place}}^{o, x, y} \gets \operatorname{PlacementQ}(\mathbf{f}_{\mathrm{place}}[o,x,y])$
    \If {$Q_{\mathrm{place}}^{o, x, y} > Q_{\mathrm{place}}^*$}
        \State $Q_{\mathrm{place}}^* \gets Q_{\mathrm{place}}^{o, x, y}$
        \State $o^*, x^*, y^* \gets o, x, y$
    \EndIf
\EndFor
\State $z^* \gets \operatorname{GetDepth}(\mathbf{H}_{\mathrm{occ}}, b^*, o^*, x^*, y^*)$
\State \Return $o^*, x^*, y^*, z^*$
\EndFunction
\end{algorithmic}
\end{algorithm}

\section{Sequential Packing Procedure}
\label{sec:sequential}

At test time, the sequential packing procedure starts with an empty container, a sequence of objects, and a buffer containing the first $|\mathcal{B}|$ unpacked objects in the sequence as candidate objects. 
A packing history list $\mathcal{H}$ is maintained to record the packed objects and their corresponding placements, which is initialized as an empty list.
The procedure consists of multiple packing steps.
At each packing step, the occupancy heightmap of the container $\mathbf{H}_{\mathrm{occ}}$ is obtained, and the candidate objects in $\mathcal{B}$ are detected.
These inputs are subsequently utilized to select the candidate object and the object placement for robotic execution.
The function \textproc{PackingStep} in Algorithm~\ref{alg:inference} summarizes the procedure of a packing step during testing.

In \textproc{PackingStep}, the algorithm first checks whether the container is full or if no candidate objects remain unpacked to determine whether the packing procedure is completed.
Otherwise, the fragility heightmap $\mathbf{H}_{\mathrm{frg}}$ is constructed from the packing history $\mathcal{H}$.
For each candidate object $b \in \mathcal{B}$, the avoidance heightmap $\mathbf{H}_{\mathrm{avd}}^b$ is constructed based on $\mathcal{H}$ and the point cloud $\mathbf{P}^b$ and property vector $\mathbf{x}_{\mathrm{pro}}^b$ are obtained.
These inputs are processed by the object encoder to produce the object feature $\mathbf{f}_{\mathrm{obj}}^b$.
The function \textproc{ChooseObject} selects the object $b^*$, and the function \textproc{ChoosePlacement} determines the placement $(o^*,x^*,y^*,z^*)$ of the object.
The resulting tuple $\mathcal{T}_{\mathrm{place}}=(b^*,o^*,x^*,y^*,z^*)$ is added to $\mathcal{H}$.

In \textproc{ChooseObject}, we set a variable $Q_{\mathrm{obj}}^*$ to record the highest Q-value for object selection. 
For each candidate object $b$, we use the object Q-predictor to obtain and compare the Q-values across all candidate objects to find $b^*$.
In \textproc{ChoosePlacement}, a variable $Q_{\mathrm{place}}^*$ is also set to record the highest Q-value for placement selection.
The candidate placement set $\mathcal{E}$ for $b^*$ is generated using the locally optimal placement selection scheme described in ``Candidate Placement Encoding'' of Section~\ref{sec:opa_net}.
Each candidate placement $(o,x,y) \in \mathcal{E}$ is parameterized by an orientation and a horizontal position.
After the optimal orientation and horizontal location $(o^*,x^*,y^*)$ are selected, the depth $z^*$ is deterministically computed as the deepest reachable location~\cite{wang2019stable} as described in ``Candidate Placement Encoding'' of Section~\ref{sec:opa_net}.
After the packing step, we update the buffer by removing the packed object $b^*$. If unpacked objects remain, the next object is added to the buffer. 

\section{Experiments}

\subsection{Implementation Details}
\label{sec:virtual_implementation}
\noindent\textbf{Datasets.} \
The virtual experiments are performed on the OPA dataset, which consists of three subsets: Kitchen (497 objects), Hardware and Stationery (302 objects), and Daily Necessities (233 objects).
The 3D models in the OPA dataset are sourced from the Objaverse-v1 dataset~\cite{deitke2023objaverse}, comprising both synthetic and scanned real-world objects.
Based on manual inspection, 40\% of the OPA objects are synthetic, 32\% are scanned from real-world objects, and 28\% are not classified into either category.
Additional details about the dataset construction are provided in Section~\ref{sec:dataset}.
We evaluate different methods on the three subsets, reporting results for each subset and averaged across all subsets.

\vspace{+5pt}
\noindent\textbf{Physical Simulation.} \
The virtual packing experiments are performed on the PyBullet physical simulator.
The container occupancy heightmap is captured using a top-down depth camera.
Following existing works~\cite{wang2019stable, zhao2023learning}, the container size is set to $32\,\mathrm{cm} \times 32\,\mathrm{cm} \times 30\,\mathrm{cm}$.
All heightmaps, including the fragility, avoidance, and occupancy heightmaps, are discretized at $1\,\mathrm{cm}$ in the $x$ and $y$ dimensions, resulting in a heightmap resolution of $32 \times 32$.
The buffer size is set to $|\mathcal{B}| = 10$, following~\cite{zhao2023learning}.
For an object $b^*$ selected by the object Q-predictor, we first extract the planar-stable poses in which $b^*$ can remain stable on a planar surface.
Across the OPA dataset, the average number of distinct planar-stable poses per object is 2.4.
For each planar-stable pose, eight in-plane rotations around the vertical z-axis are used for placement planning.
The combination of a specific planar-stable pose and an in-plane rotation forms an object orientation $o$.
For each orientation $o$, we extract all horizontal locations discretized by $2\,\mathrm{cm}$ and identify the locally optimal horizontal locations $(x, y)$ using the convex-vertex extraction method in~\cite{zhao2023learning}.
Subsequently, a maximum of $500$ candidate packing locations are employed for placement selection.

\vspace{+5pt}
\noindent\textbf{Network Training.} \
The discount factor is set to $\gamma = 0.99$ in Eq.~\eqref{eq:q_val} for deep Q-learning and the weights in Eq.~\eqref{eq:reward} are set to $\theta=15$, $\lambda=20$, and $\beta=0.2$.
In $R_{\mathrm{avd}}$, the predefined distance threshold is set to $\tau = 5\,\mathrm{cm}$, as the objects used in the experiments are sized between $5\,\mathrm{cm}$ and $30\,\mathrm{cm}$.
We set a batch size of $64$ and a learning rate of $6 \times 10^{-5}$.
The training is performed on an NVIDIA GeForce RTX 2080 GPU for 24 hours, with a maximum of 3.2 million iterations.

\begin{table}[]
  \centering
  \begin{minipage}{\linewidth}
  \caption{
  Average performance of different packing methods} over 200 random sequences. 
  OPA-Pack preserves compactness while improving avoidance accuracy and reducing the load on fragile objects.
  \label{tab:heuristic}
  \resizebox{\linewidth}{!}{
  \begin{tabular}{@{\hspace*{0mm}}l@{\hspace*{1mm}}|@{\hspace*{2mm}}c@{\hspace*{2mm}}c@{\hspace*{2mm}}c@{\hspace*{0mm}}}
    \hline
    \multicolumn{4}{c}{\textbf{Average}} \\
    \hline
    \textbf{Method}
    & \textbf{Compactness $\uparrow$} & \textbf{Avoid. Acc. $\uparrow$} & \textbf{Load on Frag. $\downarrow$ } \\
    \hline
    \textbf{OPA-Pack}  & 0.413 & \textbf{95.0}\% & 3.58\\
    \textbf{IR-BPP (Baseline)} & \textbf{0.420} & 52.3\% & 5.07\\ 
    \hline
    \textbf{MINZ}& 0.289 & 78.8\% & 4.93\\
    \textbf{HM} & 0.278 & 81.2\% & 4.86\\
    \textbf{DBL} & 0.288 & 78.8\% & 5.00\\
    \textbf{FIRSTFIT} & 0.243 & 76.3\% & \textbf{3.10}\\
    \hline
    \textbf{Rule-based MINZ} & 0.145 & 100.0\% & 0.00 \\
    \textbf{Rule-based HM} &  0.133 & 100.0\% & 0.00 \\
    \textbf{Rule-based DBL} &  0.142 & 100.0\% & 0.00 \\
    \hline 
    \hline
    \multicolumn{4}{c}{\textbf{Hardware \& Stationery}}\\
    \hline
    \textbf{Method}
    & \textbf{Compactness $\uparrow$} & \textbf{Avoid. Acc. $\uparrow$} & \textbf{Load on Frag. $\downarrow$ } \\
    \hline
    \textbf{OPA-Pack}  & 0.392 & \textbf{94.0}\% & \textbf{3.76}\\
    \textbf{IR-BPP (Baseline)} & \textbf{0.400} & 64.5\% & 6.55\\   
    \hline
    \textbf{MINZ} & 0.274 & 86.0\% & 7.13\\
    \textbf{HM} & 0.263 & 86.0\% & 6.29\\ 
    \textbf{DBL} & 0.270 & 81.0\% & 6.82\\
    \textbf{FIRSTFIT} & 0.234 & 82.5\% & 3.92\\ 
    \hline
    \textbf{Rule-based MINZ} & 0.130 & 100.0\% & 0.00 \\
    \textbf{Rule-based HM} & 0.121 & 100.0\% & 0.00 \\
    \textbf{Rule-based DBL} & 0.125 & 100.0\% & 0.00 \\
    \hline
    \hline
    \multicolumn{4}{c}{\textbf{Daily Necessities}}\\
    \hline
    \textbf{Method}
    & \textbf{Compactness $\uparrow$} & \textbf{Avoid. Acc. $\uparrow$} & \textbf{Load on Frag. $\downarrow$ } \\
    \hline
    \textbf{OPA-Pack}  & 0.415 & \textbf{94.0}\% & 3.52\\
    \textbf{IR-BPP (Baseline)} & \textbf{0.418} & 52.5\% & 4.30\\ 
    \hline
    \textbf{MINZ} & 0.258 & 79.0\% & 3.37\\
    \textbf{HM} & 0.242 & 87.0\% & 3.54\\ 
    \textbf{DBL} & 0.256 & 85.5\% & 3.43\\  
    \textbf{FIRSTFIT} & 0.226 & 77.5\% & \textbf{2.47} \\
    \hline
    \textbf{Rule-based MINZ} & 0.129 & 100.0\% & 0.00 \\
    \textbf{Rule-based HM} & 0.126 & 100.0\% & 0.00 \\
    \textbf{Rule-based DBL} & 0.126 & 100.0\% & 0.00 \\
    \hline
    \hline
    \multicolumn{4}{c}{\textbf{Kitchen}} \\
    \hline
    \textbf{Method}
    & \textbf{Compactness $\uparrow$} & \textbf{Avoid. Acc. $\uparrow$} & \textbf{Load on Frag. $\downarrow$ } \\
    \hline
    \textbf{OPA-Pack} & 0.434 & \textbf{97.0}\% & 3.45\\
    \textbf{IR-BPP (Baseline)} & \textbf{0.443} & 40.0\% & 4.36\\
    \hline
    \textbf{MINZ} & 0.335 & 71.5\% & 4.31\\
    \textbf{HM} & 0.329 & 70.5\% & 4.75\\
    \textbf{DBL} & 0.338 & 70.0\% & 4.76\\
    \textbf{FIRSTFIT}& 0.268 & 69.0\% & \textbf{2.91}\\  
    \hline
    \textbf{Rule-based MINZ} & 0.177 & 100.0\% & 0.00 \\
    \textbf{Rule-based HM} & 0.152 & 100.0\% & 0.00 \\
    \textbf{Rule-based DBL} & 0.175 & 100.0\% & 0.00 \\
    \hline
    \end{tabular}}
\end{minipage}
\end{table}

\vspace{+5pt}
\noindent\textbf{Evaluation Metrics.} \
The packing performance of different methods is evaluated using the following metrics:
\textbf{(i) Compactness}: the ratio of the total volume of packed objects to the volume of the container (Eq.~\eqref{eq:compactness}). 
\textbf{(ii) Avoidance Accuracy (Avoid. Acc.)}: a packing sequence is considered successful if no pair of already packed objects with an avoidance relation has a surface distance below $\tau = 5\,\mathrm{cm}$. 
The avoidance accuracy is computed as the fraction of successful packing sequences among all evaluated packing sequences.
\textbf{(iii) Load on Fragile Objects (Load on Frag.)}: the mean total weight exerted on each fragile object by the objects placed above it.
We report the average value across all evaluated packing sequences.
A lower value indicates reduced loading on fragile objects.

\begin{figure*}
    \centering
    \includegraphics[width=\linewidth, height=0.25\linewidth]{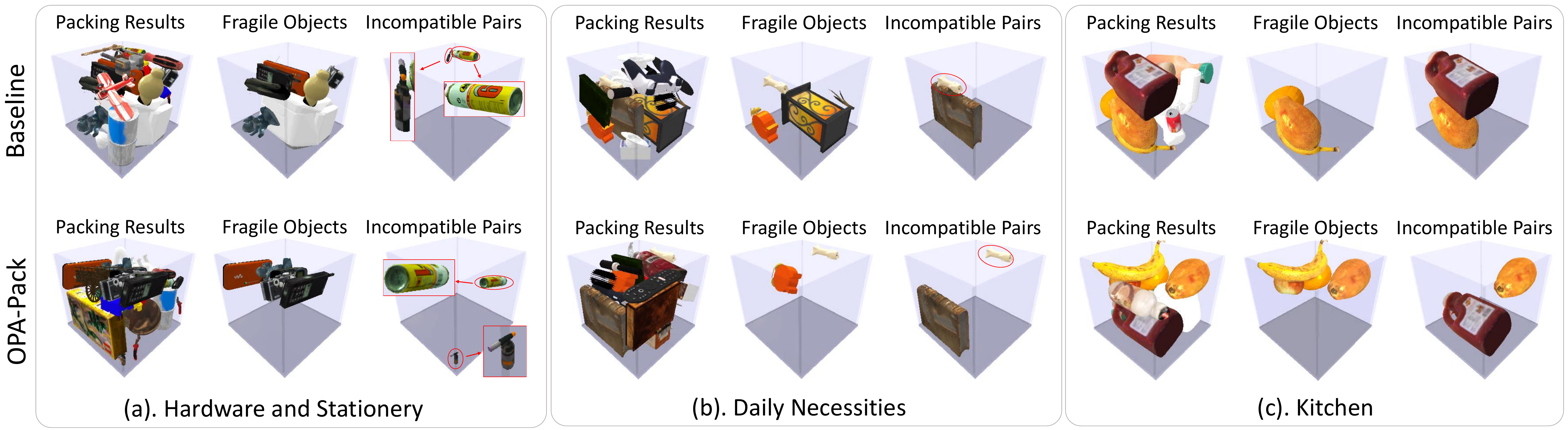}
    \caption{Visualization of the packing results produced by IR-BPP~\cite{zhao2023learning} (Baseline) and OPA-Pack from each of the following subsets: Hardware and Stationery, Daily Necessities, and Kitchen.
    In each case, the overall packing result is displayed in the first column, followed by the visualization highlighting the packed fragile objects in the second column, and a packed incompatible object pair in the last column.
    In the illustrated cases, OPA-Pack positions fragile objects in upper layers to prevent compression by heavier objects, and it results in increased separation of incompatible} object pairs.
    \label{fig:virtual_vis}
\end{figure*}

\subsection{Comparison with Existing Methods}
We perform experiments on 200 random packing sequences for each subset and compare OPA-Pack with eight existing packing methods: the IR-BPP~\cite{zhao2023learning}, a reinforcement-learning-based packing method designed to optimize packing compactness, which serves as the baseline for OPA-Pack;
Min-Z (MINZ), which prioritizes the feasible placement with the minimum placement height;
Heightmap-Minimization (HM)~\cite{wang2019stable, wang2021dense}, which prioritizes the object placement that minimizes the increase in the occupancy heightmap;
Deepest Bottom-Left (DBL)~\cite{karabulut2004hybrid}, which finds the deepest, bottom-most, and left-most position for each object; 
First-fit (FIRSTFIT), which prioritizes the first feasible object placement during search;
Rule-based MINZ, Rule-based HM, and Rule-based DBL, which are developed by modifying MINZ, HM, and DBL heuristics to include predefined deterministic rules that prohibit placing incompatible object pairs closer than $\tau$ and prevent any object from being placed on top of a fragile object.

Despite explicitly accounting for object properties, OPA-Pack achieves an average packing compactness of 0.413, which is 0.007 lower than that of IR-BPP (0.420).
Compared with IR-BPP, OPA-Pack improves the avoidance accuracy (Avoid. Acc.) from 52.3\% to 95.0\% and reduces the load on fragile objects (Load on Frag.) from 5.07 to 3.58.
OPA-Pack consistently outperforms IR-BPP on the two metrics across the three subsets.
On average, OPA-Pack achieves an avoidance accuracy of 95.0\%, compared with at most 81.2\% for FIRSTFIT, DBL, HM, and MINZ.
Although FIRSTFIT yields a reduced load on fragile objects of $3.10$, the value is associated with a low compactness of $0.243$.
Furthermore, OPA-Pack achieves higher packing compactness (0.413) compared to Rule-based DBL (0.142), Rule-based HM (0.133), and Rule-based MINZ (0.145), as the rule-based heuristics prioritize property-aware packing at the expense of compactness.

Figure~\ref{fig:virtual_vis} provides qualitative comparisons between OPA-Pack and the baseline (IR-BPP).
Results are shown for the three subsets using the same packing cases.
The qualitative results show that OPA-Pack places fragile objects in upper layers, thereby reducing the cumulative weight placed above them.
In contrast, the baseline does not explicitly model object fragility and therefore does not optimize the load placed above fragile objects. 
In addition, compared with the baseline, OPA-Pack reduces the number of incompatible object pairs packed in adjacent positions.
The improvement is attributable to explicitly modeling pairwise avoidance relations between objects during OPA-Pack policy learning.

\begin{table*}
    \centering
    \caption{Performance of the object property recognition scheme for object properties with binary results (true or false), averaged across $100$ randomly selected objects from the OPA dataset.}
    \label{tab:binary_acc}
    \begin{tabular}{c|c|c|c|c|c|c|c|c}
        \hline
          & \textbf{Soft} & \textbf{Sharp} & \textbf{Fragile} & \textbf{Edible} & \textbf{Household Chemical} & \textbf{Medicine} & \textbf{Ignition Item} & \textbf{Flammable Item} \\
        \hline
        \textbf{Accuracy} & 92\%   & 99\%   & 90\%    & 94\% & 99\%  & 98\%  & 100\%  & 93\% \\
        \hline
        \textbf{Precision} & 89\%   & 100\%  & 96\%    & 96\% & 100\% & 85\%  & 100\%  & 85\% \\
        \hline
        \textbf{Recall} & 89\%   & 86\%   & 86\%    & 92\% & 50\% & 85\%  & 100\%  & 80\% \\
        \hline
        \textbf{F1-score} & 89\%   & 92\%   & 90\%    & 94\% & 67\% & 85\%  & 100\%  & 83\% \\
        \hline
    \end{tabular}
\end{table*}

\begin{table}[]
    \centering
    \caption{Performance of the object property recognition scheme for object properties with non-binary results, averaged across $100$ randomly selected objects from the OPA dataset.}
    \label{tab:non_binary_acc}
    \begin{tabular}{c|c|c|c}
        \hline
         & \textbf{Material In} & \textbf{Material Out} & \textbf{Density Level} \\
         \hline
         \textbf{Accuracy} & 99\%   & 95\%     & 93\%  \\
        \hline
    \end{tabular}
\end{table}

\begin{figure*} [b]
    \centering
    \includegraphics[width=\linewidth]{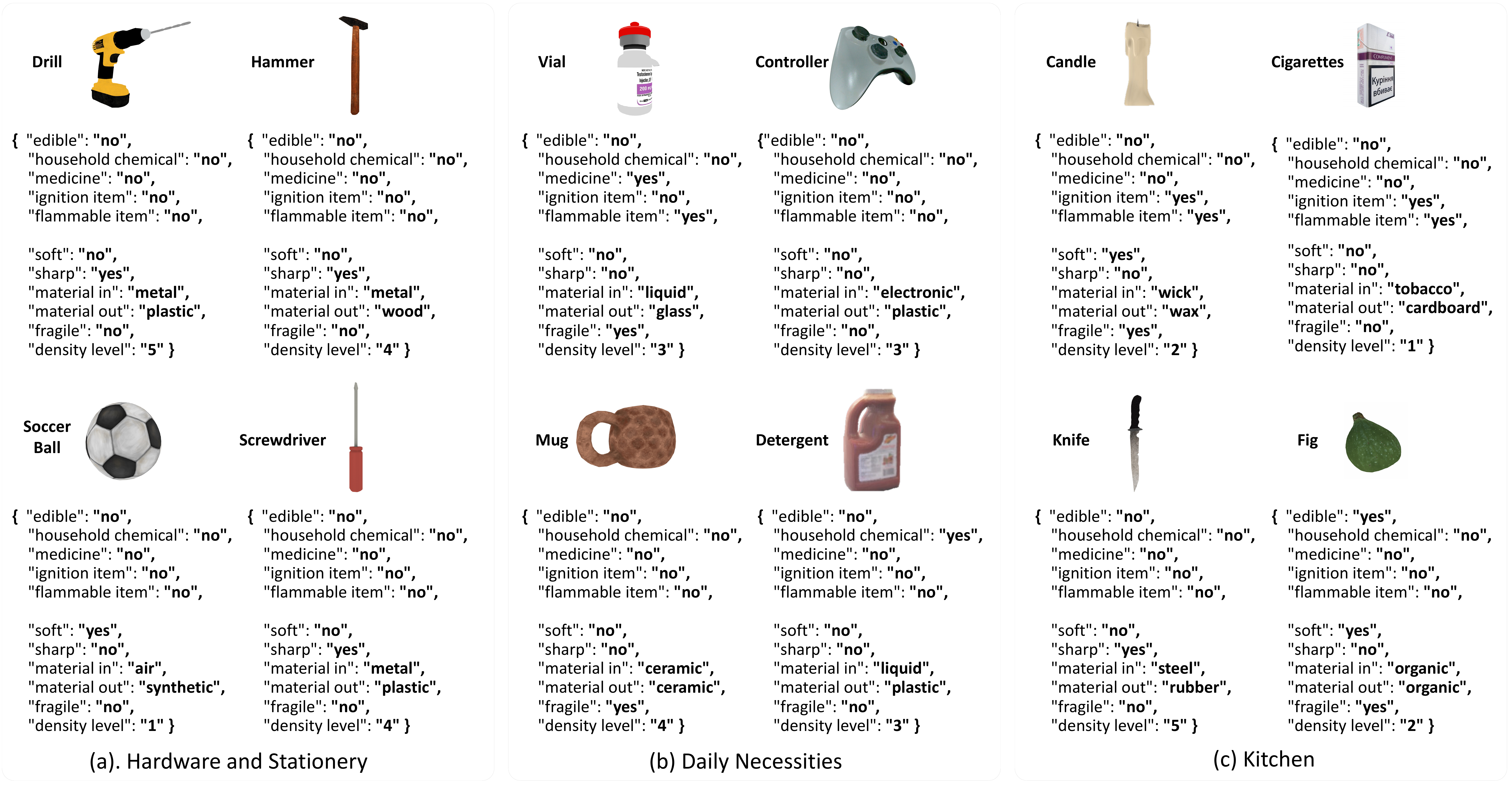}
    \caption{Visualization of the property recognition results in the OPA dataset. OPA-Pack predicts detailed object properties, including semantic properties (\ie, edibility, household chemicals, medicines, ignition items, and flammable items) and physical attributes (\ie, softness, sharpness, inner and outer materials, fragility, and density levels). 
    }
    \label{fig:property_vis}
\end{figure*}

\subsection{Object Property Recognition Results}
We randomly select 100 objects from the OPA dataset for the evaluation.
Table~\ref{tab:binary_acc} reports the accuracy, precision, recall, and F1-score for each object property with a binary result (true or false) predicted by the object property recognition scheme in OPA-Pack.
Table~\ref{tab:non_binary_acc} reports the accuracy of non-binary object properties.
Across the evaluated object properties, accuracy ranges from $90\%$ to $100\%$.
For all binary properties except ``Household Chemicals'', the precision, recall, and F1-score are at least $85\%$, $80\%$, and $83\%$, respectively, while predicting the household-chemical property remains challenging for the vision-language model.
Sections~\ref{sec:ablation} and~\ref{sec:parameter} present the ablation studies and additional statistical analyses.

Figure~\ref{fig:property_vis} visualizes the recognition results for four objects from each subset of the OPA dataset.
Each object is represented by its type name from the Objaverse dataset~\cite{deitke2023objaverse}, five semantic properties (\ie, edibility, household chemicals, medicines, ignition items, and flammable items) and six physical properties (\ie, softness, sharpness, inner material, outer material, fragility, and density level).
The object property recognition scheme identifies the fig as edible, the detergent as a household chemical, the vial as medicine, and the candle and cigarettes as ignition items.
The vial, candle, and cigarettes are predicted as flammable items.
For physical properties, the proposed method recognizes vial, mug, candle, and fig as fragile objects; the soccer ball, candle, and fig as soft objects; and the drill, hammer, screwdriver, and knife as sharp objects.
The inner and outer materials of the objects are also predicted.
For single-material objects, such as the ceramic mug and organic fig, the predicted inner and outer materials are identical.
For multi-material objects, the predicted inner and outer materials differ: the drill and screwdriver are predicted as plastic externally and metal internally; the vial as glass externally and liquid internally; and the cigarettes as cardboard externally and tobacco internally.
For density level, the drill and knife are assigned a density level of five; the detergent and vial are assigned a density level of three; and the soccer ball and cigarettes are assigned a density level of one.

\subsection{Ablation Studies}
\label{sec:ablation}
Ablation studies are performed to evaluate the contributions of the major designs in the proposed approach.
We evaluate object property recognition by component removal and object-property-aware packing learning through cumulative additions of components to the baseline.

\begin{figure*}
    \centering
    \includegraphics[width=\linewidth]{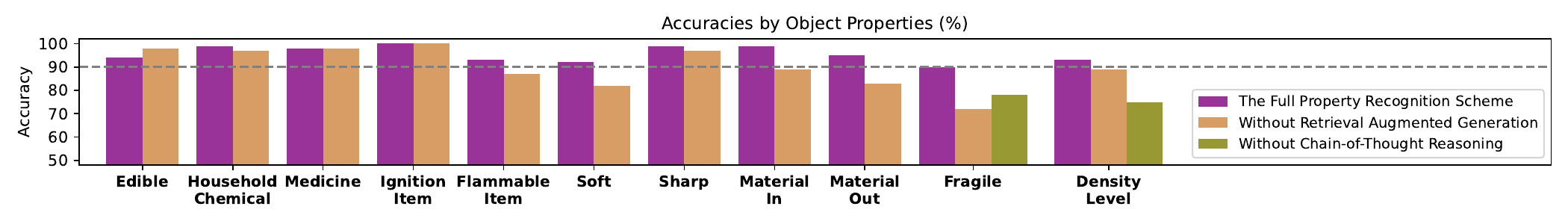}
    \caption{The ablation studies of the design components in the object property recognition scheme.
    The annotation accuracies are evaluated using 100 randomly selected objects from the OPA dataset.
    Retrieval-augmented generation and chain-of-thought reasoning improve the accuracy of object property recognition.
    }
    \label{fig:llm_ablation}
\end{figure*}

\begin{table*}
  \centering
  \begin{minipage}{\linewidth}
  \caption{Ablation studies of OPA-Net. The fragility reward term, the avoidance reward term, and the network architecture contribute to improving the packing compactness, increasing the avoidance accuracy, and reducing the load on fragile objects. \vspace{-5pt}}
  \label{tab:ablation}
  \resizebox{\linewidth}{!}{
  \begin{tabular}{@{\hspace*{2mm}}c@{\hspace*{2mm}}|c@{\hspace*{2mm}}c@{\hspace*{2mm}}c@{\hspace*{2mm}}c@{\hspace*{2mm}}|@{\hspace*{2mm}}c@{\hspace*{2mm}}c@{\hspace*{2mm}}c@{\hspace*{0mm}}}
    \hline
      &\textbf{$R_{\mathrm{frg}}$} & \textbf{$R_{\mathrm{avd}}$} & \textbf{Property Embedding} & \textbf{Map Encoders} & \textbf{Compactness $\uparrow$} & \textbf{Avoid. Acc. $\uparrow$} & \textbf{Load on Frag. $\downarrow$} \\
    \hline
   \multirow{5}{*}{\textbf{Average}} &  &  &  &  & \textbf{0.420} & 52.33\% & 5.07\\
    & \ding{52} &  &  &  & 0.419 & 83.33\% & 4.45\\
    & \ding{52} & \ding{52} &  &  & 0.378 & 91.50\% & 4.23\\
    & \ding{52} & \ding{52} & \ding{52} &  &  0.412 & 92.50\%  & 3.86 \\
    & \ding{52} & \ding{52} & \ding{52} &  \ding{52} & 0.413 & \textbf{95.00}\% & \textbf{3.58}\\
    \hline
    \multirow{5}{*}{\textbf{Hardware \& Stationery}} &  &  &  &  & \textbf{0.400} & 64.50\% & 6.55 \\
    & \ding{52} &  &  &  & 0.384 & 82.00\% & 4.73 \\
    & \ding{52} & \ding{52} &  &  & 0.371 & 89.50\% & 4.68\\
    & \ding{52} & \ding{52} & \ding{52} &  &  0.391 & 91.50\% & 4.26 \\
    & \ding{52} & \ding{52} & \ding{52} & \ding{52} & 0.392 & \textbf{94.00}\% & \textbf{3.76} \\ 
    \hline
    \multirow{5}{*}{\textbf{Daily Necessities}} &  &  &  &  & 0.418 & 52.50\% & 4.30 \\
    & \ding{52} &  &  &  & \textbf{0.439} & 86.50\% & 4.89\\
    & \ding{52} & \ding{52} &  &  & 0.408 & 91.00\% & 4.24\\
    & \ding{52} & \ding{52} & \ding{52} &  &  0.406 & \textbf{94.00}\%  & 3.76 \\
    & \ding{52} & \ding{52} & \ding{52} & \ding{52} & 0.415 & \textbf{94.00}\% & \textbf{3.52}\\
    \hline
    \multirow{5}{*}{\textbf{Kitchen}} &  &  &  &  & \textbf{0.443} & 40.00\% & 4.36\\
    & \ding{52} &  &  &  & 0.435 & 81.50\% & 3.72 \\
    & \ding{52} & \ding{52} &  &  & 0.372 & 94.00\% & \textbf{3.40} \\
    & \ding{52} & \ding{52} & \ding{52} &  &  0.440 & 92.00\%  & 3.56 \\
    & \ding{52} & \ding{52} & \ding{52} & \ding{52} &  0.434 & \textbf{97.00}\% & 3.45 \\
    \hline
    \end{tabular}
    }
\end{minipage}
\end{table*}

\vspace{+5pt}
\noindent\textbf{Object Property Recognition.} \
The ablation studies on object property recognition evaluate two designs: 
(i) retrieval-augmented generation, which uses an auxiliary database with human annotations to provide retrieved examples for property prediction; and 
(ii) chain-of-thought reasoning, which predicts the object's inner and outer materials before estimating fragility and density level.
To assess the annotation quality of different designs, 100 objects are randomly selected from the OPA dataset, and a human participant evaluates whether each property prediction is correct.

The results are reported in Figure~\ref{fig:llm_ablation}. 
Removing retrieval-augmented generation reduces recognition accuracy for household chemicals, flammable items, softness, sharpness, inner and outer materials, fragility, and density level, compared with the full property recognition scheme.
Removing chain-of-thought reasoning results in consistent reductions in the recognition accuracy of fragility and density level.
These results indicate that both retrieval-augmented generation and chain-of-thought reasoning contribute to object property recognition.

\vspace{+5pt}

\noindent\textbf{Object-Property-Aware Packing.} \
Additional ablation studies evaluate each component of object-property-aware packing learning.
The results are shown in Table~\ref{tab:ablation}.
The baseline contains only the components required to optimize packing compactness: the object geometric encoder $E_{\mathrm{pcl}}(\cdot)$, the candidate pose-embedding layer $E_{\mathrm{pos}}(\cdot)$, the container's occupancy heightmap $\mathbf{H}_{\mathrm{occ}}$, the object Q-predictor, the placement Q-predictor, and the compactness reward in Eq.~\eqref{eq:compactness}.
The fragility reward term $R_{\mathrm{frg}}$ and the avoidance reward term $R_{\mathrm{avd}}$ are subsequently included to penalize load on fragile objects and separate incompatible object pairs.
The property-embedding layer $E_{\mathrm{pro}}(\cdot)$, denoted as ``Property Embedding'' in Table~\ref{tab:ablation}, is also included to encode the properties of candidate objects.
The encoders for fragility heightmap $\mathbf{H}_{\mathrm{frg}}$ and avoidance heightmap $\mathbf{H}_{\mathrm{avd}}$, denoted as ``Map Encoders'' in Table~\ref{tab:ablation}, are further integrated for encoding the locations of previously packed fragile and incompatible objects.

The results show that incorporating the fragility reward term reduces the average load on fragile objects from 5.07 to 4.45.
Further incorporating the avoidance reward term increases avoidance accuracy from 83.33\% to 91.50\%.
These results demonstrate the contribution of the reward design.
Including the property-embedding layer and the map encoders for the fragility and avoidance heightmaps increases packing compactness from 0.378 to 0.413 and avoidance accuracy from 91.50\% to 95.00\%, while reducing the load on fragile objects from 4.23 to 3.58.

Furthermore, we evaluate the impact of introducing candidate poses as inputs to the object Q-predictor.
For each candidate object $b$ in the buffer $\mathcal{B}$, candidate orientations $o$ and horizontal placement locations $(x,y)$ are pre-computed.
The corresponding pose features are extracted and aggregated using max pooling to obtain an object-level pose representation, which is used as input to the object Q-predictor.
As shown in Table~\ref{tab:pose_to_object_q}, incorporating candidate pose features increases packing compactness from 0.413 to 0.470, while decreasing avoidance accuracy from 95.0\% to 91.5\% and increasing the load on fragile objects from 3.58 to 3.79.
In addition, candidate-pose extraction requires 0.41 seconds per object on average, whereas network inference requires 0.02 seconds.
Pre-computing poses for all candidate objects therefore introduces computational overhead without yielding consistent performance improvements.
Based on these observations, the candidate pose features are not incorporated in the object Q-predictor.

\begin{table}[t] 
  \centering
  \begin{minipage}{\linewidth}
  \caption{Ablation analysis of including candidate poses in the object Q-predictor. \vspace{-5pt}}
  \label{tab:pose_to_object_q}
  \resizebox{\linewidth}{!}{
  \begin{tabular}{@{\hspace*{0mm}}l@{\hspace*{1mm}}|@{\hspace*{2mm}}c@{\hspace*{2mm}}c@{\hspace*{2mm}}c@{\hspace*{2mm}}c@{\hspace*{0mm}}}
    \hline
    & \textbf{Compactness $\uparrow$} & \textbf{Avoid. Acc. $\uparrow$} & \textbf{Load on Frag. $\downarrow$ }\\
    \hline
    \textbf{With Candidate Poses} & \textbf{0.470} & 91.5\% & 3.79\\ 
    \textbf{Without Candidate Poses}  & 0.413 & \textbf{95.0}\% & \textbf{3.58}\\
    \hline 
    \end{tabular}
    }
\end{minipage}
\end{table}

\subsection{Additional Analysis}
\label{sec:parameter}
This section provides additional analyses of the object property recognition scheme and object-property-aware packing learning under different experimental settings.

\vspace{+5pt}
\noindent\textbf{Stability of Object Property Recognition across Multiple Runs.}
The experiment is conducted on 100 objects randomly selected from the OPA dataset. 
The object property recognition scheme is run ten times, and the accuracy of each object property is reported.
For each run, the recognition accuracy is averaged over the 100 objects. 
The mean and standard deviation (Std. Dev.) of each object property are computed across the ten runs.
Table~\ref{tab:property_multiple_runs} shows that the recognition accuracy remains at least $89.1\%$ across properties, with a standard deviation of at most $2.71$ across runs.
The cross-run prediction agreement over 10 runs is additionally reported in the supplementary material.

\begin{table}[]
    \centering
    \caption{Mean and standard deviation (std. dev.) of object property recognition results obtained across ten runs.\vspace{-5pt}}
    \begin{tabular}{c|c|c}
        \hline
           & \textbf{Mean Accuracy} $\uparrow$ & \textbf{Std. Dev.} $\downarrow$ \\ 
        \hline
        \textbf{Edible}          & 91.9\%         & 2.70               \\
        \textbf{Household Chemical} & 98.8\%      & 0.40               \\
        \textbf{Medicine}        & 97.1\%         & 0.83               \\ 
        \textbf{Ignition Item}   & 99.8\%         & 0.40               \\ 
        \textbf{Flammable Item}  & 92.4\%         & 0.92               \\ 
        \textbf{Softness}        & 92.8\%         & 1.25               \\ 
        \textbf{Sharpness}       & 98.3\%         & 0.64               \\ 
        \textbf{Material In}     & 97.6\%         & 1.28               \\ 
        \textbf{Material Out}    & 94.7\%         & 1.00               \\
        \textbf{Fragile}         & 89.1\%         & 1.45               \\
        \textbf{Density Level}   & 92.8\%         & 2.71               \\
        \hline
    \end{tabular}
    \label{tab:property_multiple_runs}
\end{table}

\begin{figure*}
    \centering
    \includegraphics[width=\linewidth]{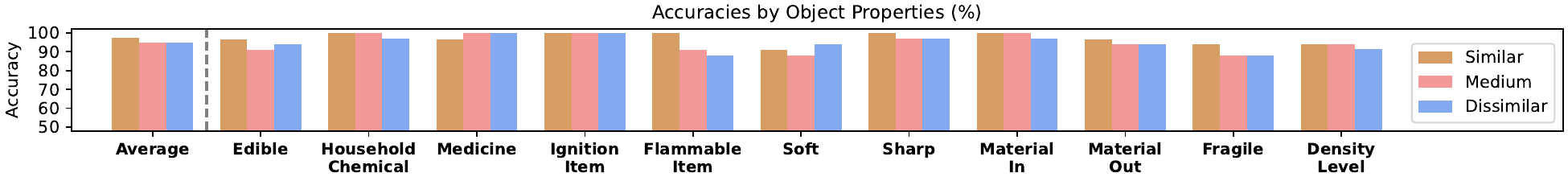}
    \caption{The property recognition results of objects with varying similarity levels to the auxiliary database, \ie, similar, medium, and dissimilar.
    The first column reports the mean accuracies across all eleven object properties for each similarity level, followed by the per-property accuracies.
    }
    \label{fig:llm_similarity}
\end{figure*}

\vspace{+5pt}
\noindent\textbf{Property Recognition for Objects with Varying Similarity Levels to the Auxiliary Database.}
The object property recognition scheme is evaluated on objects with different levels of similarity to the auxiliary database $\mathcal{U}_{\mathrm{aux}}$.
For each object $u$ from the OPA dataset and each auxiliary object $u' \in \mathcal{U}_{\mathrm{aux}}$, their front-view CLIP features are extracted, and their cosine similarity is computed.
The similarity score of object $u$ to the auxiliary database $\mathcal{U}_{\mathrm{aux}}$ is defined as
$S_{\mathrm{sim}}(u) = \max_{u' \in \mathcal{U}_{\mathrm{aux}}} \operatorname{cosine}(E_{\mathrm{CLIP}}(u), E_{\mathrm{CLIP}}(u'))$, where $\operatorname{cosine}(\cdot, \cdot)$ denotes the cosine similarity, and $E_{\mathrm{CLIP}}(\cdot)$ denotes the CLIP encoder applied to the front-view image of each object.
The objects from the OPA dataset are evenly partitioned into three groups based on $S_{\mathrm{sim}}(u)$: similar, medium-similarity, and dissimilar.
For each group, $33$ objects are randomly sampled, and the mean recognition accuracy for each object property is computed.
Figure~\ref{fig:llm_similarity} shows that objects in the ``similar'' group have higher mean accuracy than those in the ``medium'' and ``dissimilar'' groups.
The recognition accuracy remains above 85\% for all three similarity levels.
Results with six-level similarity partitions are shown in the supplementary material.

\begin{figure*}
    \centering
    \includegraphics[width=0.98\linewidth]{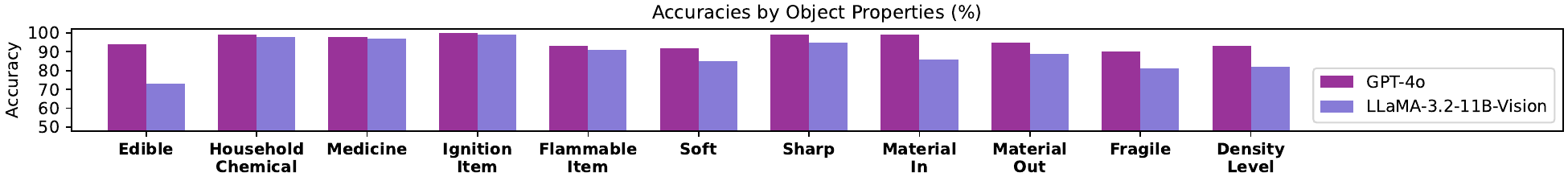}
    \caption{
    Performance of object property recognition using GPT-4o~\cite{achiam2023gpt} and using LLaMA-3.2-11B-Vision~\cite{dubey2024llama}.
    }
    \label{fig:llm_gpt_llama}
\end{figure*}

\vspace{+5pt}
\noindent\textbf{Object Property Recognition using Different VLMs.}
GPT-4o~\cite{achiam2023gpt} and LLaMA-3.2-11B-Vision~\cite{dubey2024llama} are evaluated for object property recognition on 100 randomly selected objects from the OPA dataset.
As shown in Figure~\ref{fig:llm_gpt_llama}, GPT-4o~\cite{achiam2023gpt} consistently outperforms LLaMA-3.2-11B-Vision~\cite{dubey2024llama} in predicting object properties. 
GPT-4o~\cite{achiam2023gpt} is therefore used in the main experiments.

\begin{figure}
    \centering
    \includegraphics[width=\linewidth]{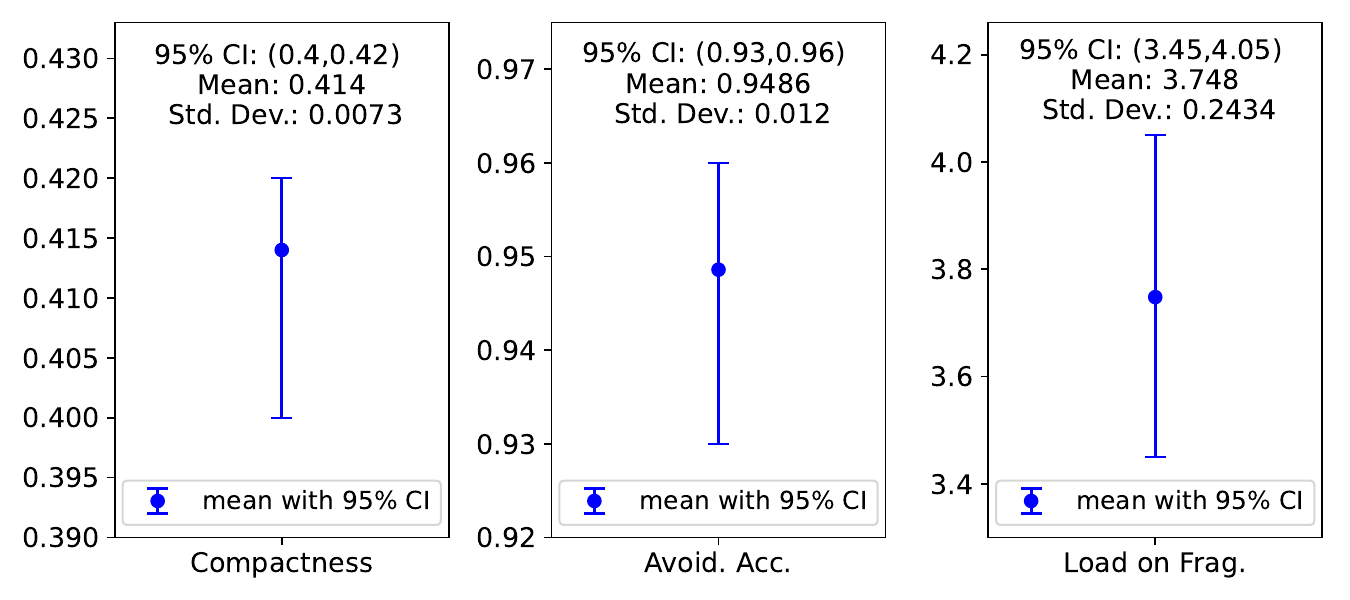}
    \caption{Performance of OPA-Net on packing compactness, avoidance accuracy (Avoid. Acc.), and load on fragile objects (Load on Frag.) across five random seeds. 
    For each metric, we show the 95\% confidence intervals (CIs), mean, and standard deviation (Std. Dev.).}
    \label{fig:network_confidence}
\end{figure}

\vspace{+5pt}
\noindent\textbf{Stability of OPA-Net across Multiple Random Seeds.}
We train OPA-Net with five different random seeds and report the compactness, avoidance accuracy (Avoid. Acc.), and load on fragile objects (Load on Frag.) of the five resulting models in Figure~\ref{fig:network_confidence}.
For each metric, the 95\% confidence interval (CI), the mean, and the standard deviation (Std. Dev.) across the five models are reported in Figure~\ref{fig:network_confidence}.
The results show that the 95\% CI width for each metric is $0.02$ for compactness, $0.03$ for avoidance accuracy, and $0.6$ for load on fragile objects.
The standard deviation is at most $0.2434$ for all three evaluated metrics.
Across the five evaluated random seeds, OPA-Net shows limited variation in the three reported metrics.

\vspace{+5pt}
\noindent\textbf{OPA-Net under Different Buffer Sizes.}
OPA-Net is evaluated with different buffer sizes $|\mathcal{B}|$: $|\mathcal{B}| = 5$, $|\mathcal{B}| = 10$, and $|\mathcal{B}| = 15$.
Results are presented in Table~\ref{tab:buffer_sizes}.
In the main experiments, we use a buffer size of $|\mathcal{B}| = 10$.
Increasing the buffer size from $5$ to $15$ improves packing compactness from 0.348 to 0.450 and avoidance accuracy from 91.5\% to 96.3\%.
Compared with $|\mathcal{B}| = 10$, the load on fragile objects for $|\mathcal{B}| = 15$ increases from $3.58$ to $3.78$.
$|\mathcal{B}| = 10$ is therefore used in the main experiments.

\begin{table}
  \centering
  \begin{minipage}{0.9\linewidth}
  \caption{The performance of OPA-Net under buffer sizes $|\mathcal{B}| = 5$, $|\mathcal{B}| = 10$, and $|\mathcal{B}| = 15$. The average results across the three subsets of the OPA dataset are presented.}
  \label{tab:buffer_sizes}
  \resizebox{\linewidth}{!}{
  \begin{tabular}{@{\hspace*{2mm}}c@{\hspace*{2mm}}|@{\hspace*{2mm}}c@{\hspace*{2mm}}c@{\hspace*{2mm}}c@{\hspace*{0mm}}}
   \hline
    \textbf{Buffer Size $|\mathcal{B}|$}  &  \textbf{Compactness $\uparrow$} & \textbf{Avoid. Acc. $\uparrow$} & \textbf{Load on Frag. $\downarrow$}\\
    \hline
    \textbf{$|\mathcal{B}| = 5$} &  0.348  & 91.5\% &  3.83 \\
   \textbf{$|\mathcal{B}| = 10$} & 0.413  & 95.0\% &  3.58 \\   
   \textbf{$|\mathcal{B}| = 15$}& 0.450  & 96.3\%  &  3.78 \\ 
    \hline
\end{tabular}}
\end{minipage}
\end{table}

\vspace{+5pt}
\noindent\textbf{OPA-Net under Different Container Sizes.}
We train and evaluate OPA-Net using three different container sizes, $24\,\mathrm{cm} \times 24\,\mathrm{cm}$, $32\,\mathrm{cm} \times 32\,\mathrm{cm}$, and $48\,\mathrm{cm} \times 48\,\mathrm{cm}$, each with a height of $30\,\mathrm{cm}$.
Average results for packing compactness, avoidance accuracy, and load on fragile objects in the OPA dataset are reported.
Results in Table~\ref{tab:container_sizes} show that reducing the container size from $32\,\mathrm{cm} \times 32\,\mathrm{cm}$ to $24\,\mathrm{cm} \times 24\,\mathrm{cm}$ changes the packing compactness from 0.413 to 0.403 and the avoidance accuracy from 95.0\% to 93.0\%, while reducing the load on fragile objects from 3.58 to 2.96.
For the larger container size ($48\,\mathrm{cm} \times 48\,\mathrm{cm}$), compactness decreases to 0.325, avoidance accuracy decreases to 90.8\%, and load on fragile objects increases to 3.87.
One possible explanation for the performance degradation is that the enlarged search space of candidate packing locations increases the optimization complexity.

\begin{table}
  \centering
  \begin{minipage}{0.9\linewidth}
  \caption{The performance of OPA-Net under different container sizes, averaged across the three OPA dataset subsets. OPA-Net is evaluated on three sizes: $24\,\mathrm{cm} \times 24\,\mathrm{cm}$, $32\,\mathrm{cm} \times 32\,\mathrm{cm}$, and $48\,\mathrm{cm} \times 48\,\mathrm{cm}$, all with $30\,\mathrm{cm}$ in height.}
  \label{tab:container_sizes}
  \resizebox{\linewidth}{!}{
  \begin{tabular}{@{\hspace*{2mm}}c@{\hspace*{2mm}}|@{\hspace*{2mm}}c@{\hspace*{2mm}}c@{\hspace*{2mm}}c@{\hspace*{0mm}}}
   \hline
     \textbf{Container Sizes} &  \textbf{Compactness $\uparrow$} & \textbf{Avoid. Acc. $\uparrow$} & \textbf{Load on Frag. $\downarrow$}\\
    \hline
    \textbf{$24\,\mathrm{cm} \times 24\,\mathrm{cm}$} &  0.403  & 93.0\% &  2.96 \\
   \textbf{$32\,\mathrm{cm} \times 32\,\mathrm{cm}$} & 0.413  & 95.0\% &  3.58 \\   
   \textbf{$48\,\mathrm{cm} \times 48\,\mathrm{cm}$}& 0.325  & 90.8\%  &  3.87 \\ 
    \hline
\end{tabular}}
\end{minipage}
\end{table}

\vspace{+5pt}
\noindent\textbf{OPA-Net under Different Weights of the Reward Terms.}
A parameter analysis is conducted for $\lambda$ and $\beta$ in the reward design (Eq.~\eqref{eq:reward}) by varying each parameter while keeping the other fixed.
$\lambda$ represents the weight of the fragility reward term, while $\beta$ denotes the weight of the avoidance reward term.
A larger $\lambda$ increases the penalty for placing objects on top of a fragile object, whereas a larger $\beta$ increases the penalty for placing incompatible object pairs in close proximity.
Each parameter is independently scaled up and down by a factor of $10$, and the corresponding metrics are reported in Table~\ref{tab:reward_weights}.

\begin{table}
  \centering
  \begin{minipage}{0.9\linewidth}
  \caption{OPA-Net under different weights of the reward terms. The average results across the three subsets of the OPA dataset are reported.}
  \label{tab:reward_weights}
  \resizebox{\linewidth}{!}{
  \begin{tabular}{@{\hspace*{2mm}}c@{\hspace*{2mm}}|@{\hspace*{2mm}}c@{\hspace*{2mm}}c@{\hspace*{2mm}}c@{\hspace*{0mm}}}
   \hline
    \multicolumn{4}{c}{}\\
    \hline
      &  \textbf{Load on Frag. $\downarrow$} & \textbf{Compactness $\uparrow$} &  \textbf{Avoid. Acc. $\uparrow$}\\
    \hline
    \textbf{$\lambda$ = 2} &  4.38  & 0.423 &  91.00\% \\ 
   \textbf{$\lambda$ = 20} & 3.58  & 0.413 &  95.00\% \\   
   \textbf{$\lambda$ = 200}& 3.51  & 0.344  &  92.33\% \\ 
    \hline
    \hline
    \multicolumn{4}{c}{}\\
    \hline
      & \textbf{Avoid. Acc. $\uparrow$} & \textbf{Compactness $\uparrow$} & \textbf{Load on Frag. $\downarrow$} \\
    \hline
    \textbf{$\beta$ = 0.02} &  76.33\% & 0.424 &  3.32 \\  
   \textbf{$\beta$ = 0.2} & 95.00\%  & 0.413 &  3.58 \\    
   \textbf{$\beta$ = 2}&  96.50\% &  0.378 &  3.29 \\ 
    \hline
\end{tabular}}
\end{minipage}
\end{table}

Decreasing $\lambda$ to $2$ results in an increase of $22\%$ in load on fragile objects.
Compared with $\lambda = 20$, increasing $\lambda$ to $200$ reduces the load on fragile objects by $2\%$. 
However, this setting decreases packing compactness by $0.069$. 
Therefore, $\lambda = 20$ is adopted to balance packing compactness and load on fragile objects.
Decreasing $\beta$ from $0.2$ to $0.02$ reduces avoidance accuracy from 95.00\% to 76.33\%.
Increasing $\beta$ from $0.2$ to $2$ improves avoidance accuracy from 95.00\% to 96.50\% but decreases packing compactness from 0.413 to 0.378.
Therefore, $\beta = 0.2$ is adopted in the main experiments.

\subsection{Evaluation on Scanned Real-World Objects}

\label{sec:real_virtual}

\begin{table}[t] 
  \centering
  \begin{minipage}{\linewidth}
  \caption{Comparing the average performance of OPA-Pack and the baseline over 200 random sequences on the scanned real-world objects in the physical simulator.
  Compared with the baseline, OPA-Pack achieves higher avoidance accuracy and reduces the load on fragile objects while preserving
  similar compactness.
  }
  \label{tab:real_virtual}
  \resizebox{\linewidth}{!}{
  \begin{tabular}{@{\hspace*{0mm}}l@{\hspace*{1mm}}|@{\hspace*{2mm}}c@{\hspace*{2mm}}c@{\hspace*{2mm}}c@{\hspace*{0mm}}}
    \hline
    \multicolumn{4}{c}{\textbf{Scanned Real-World Objects}}\\
    \hline
    \textbf{Method}
    & \textbf{Compactness $\uparrow$} & \textbf{Avoid. Acc. $\uparrow$} & \textbf{Load on Frag. $\downarrow$ } \\
    \hline
    \textbf{OPA-Pack} & \textbf{0.621} & \textbf{93.0}\% & \textbf{1.01} \\
    \textbf{IR-BPP (Baseline)} & 0.616 & 51.0\% & 1.10 \\  
    \hline
    \textbf{MINZ} & 0.521 & 51.5\% & 1.13 \\  
    \textbf{HM} & 0.503 & 59.5\% &  1.10\\  
    \textbf{DBL} & 0.510 & 56.0\% & 1.19 \\  
    \hline
    \end{tabular}
    }
\end{minipage}
\end{table}

To evaluate applicability beyond the OPA dataset, a dataset comprising $100$ scanned real-world objects is collected.
For each object, a 3D CAD model is reconstructed using the Scan3D App, and object property recognition is performed following the procedure described in Section~\ref{sec:dataset}.
The reconstructed objects are loaded into the virtual environment for packing evaluation.
OPA-Pack is evaluated in comparison with IR-BPP~\cite{zhao2023learning} and three packing heuristics: MINZ, HM~\cite{wang2019stable, wang2021dense}, and DBL~\cite{karabulut2004hybrid}.
The results are shown in Table~\ref{tab:real_virtual}.

OPA-Pack achieves a packing compactness of 0.621, which is 0.005 higher than 0.616 achieved by IR-BPP.
It also achieves an avoidance accuracy of $93.0\%$, outperforming the 51.0\% achieved by IR-BPP.
OPA-Pack also reduces the load on fragile objects from 1.10 to 1.01 compared with IR-BPP.
These results demonstrate the applicability of OPA-Pack to scanned real-world objects.

\section{Robotic Packing on Physical Platform}
\subsection{Platform Setup}
\label{sec:platform}
\begin{figure}
    \centering
    \includegraphics[width=0.86\linewidth]{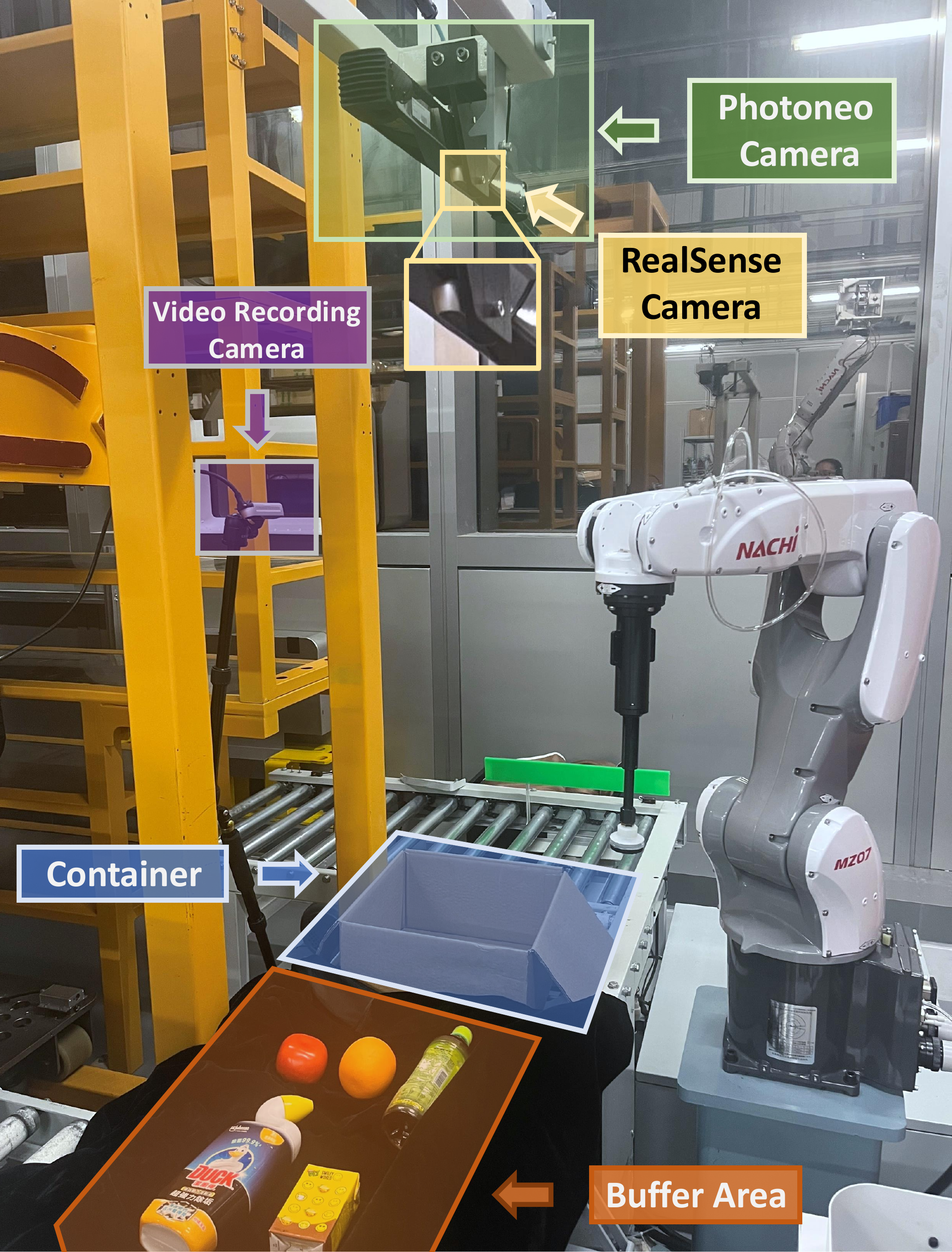}
    \caption{The physical platform with a NACHI MZ07 robot arm. A buffer area (orange, bottom) is set to store five candidate objects; a container (blue, center) is used for packing objects; a Photoneo camera (green, top) is used for depth sensing; a RealSense (yellow, top) camera is utilized to capture the RGB image; and another camera (purple, top-left) is used for video recording.
    }
    \label{fig:real_platform}
\end{figure}

\noindent\textbf{Hardware and Network Settings.} \
The physical platform consists of a NACHI MZ07 robot equipped with a suction cup for object manipulation, as well as a Photoneo camera and a RealSense camera for acquiring top-down observations of the container and buffer area.
The buffer area consists of five regions, each of which stores one object.
The buffer area is manually refilled after all objects have been placed.
The container size is $30\,\mathrm{cm} \times 30\,\mathrm{cm} \times 20\,\mathrm{cm}$, and the buffer size is $|\mathcal{B}| = 5$.
OPA-Net is trained using the $100$ scanned real-world objects described in Section~\ref{sec:real_virtual}, and evaluated on the physical platform.
Unless otherwise stated, the settings from Section~\ref{sec:virtual_implementation} are used: heightmaps are discretized at a resolution of $1\,\mathrm{cm}$ along both the $x$ and $y$ dimensions; object orientations are discretized into eight in-plane rotations; horizontal placement locations are discretized at a resolution of $2\,\mathrm{cm}$; the reward weights are $\theta=15$, $\lambda=20$, and $\beta=0.2$; the batch size is $64$; and the learning rate is $6 \times 10^{-5}$.
Figure~\ref{fig:real_platform} illustrates the physical platform.

\vspace{+5pt}

\noindent\textbf{Packing Pipeline.} \
A real-world packing sequence consists of multiple packing steps. 
Each step involves selecting an object and placing it into the container.
On the physical platform, each packing step consists of four stages: (1) object identification and candidate-object selection, (2) initial-pose estimation, (3) placement-pose determination in the container, and (4) robotic pick-and-place action generation and execution.
The packing pipeline also includes (5) an error recovery mechanism for detecting and recovering from picking failures and sensing errors.
The five components are described below.

\begin{figure*}
    \centering
    \includegraphics[width=0.89\linewidth, height=0.63\linewidth]{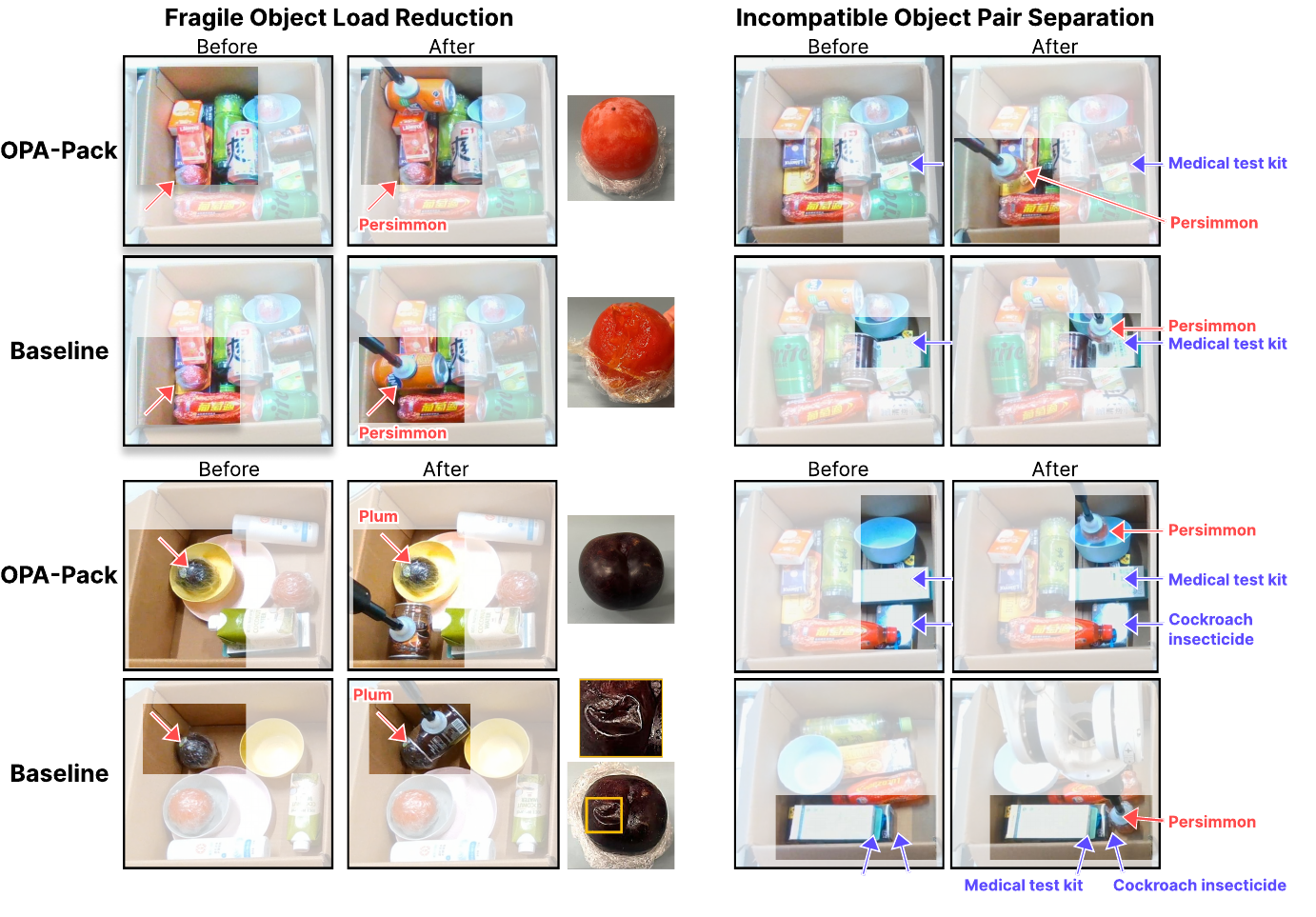}
    \caption{Visual comparison of OPA-Pack and IR-BPP~\cite{zhao2023learning} (Baseline) based on real-world packing results. 
    Left: Compared with the baseline, OPA-Pack reduces the placement of heavy objects on fragile persimmons and plums, thereby reducing potential fruit damage.
    Right: The baseline packs the persimmon adjacent to the medical test kit or the cockroach insecticide, whereas OPA-Pack maintains separation between these incompatible objects.
    }
    \label{fig:real_vis}
\end{figure*}

\vspace{+5pt}
\noindent\textbf{Object Identification and Selection.} \
Objects in the buffer area are identified based on the captured point cloud.
For each buffer region corresponding to a candidate object, a predefined 3D bounding box is constructed to crop the local point cloud.
A region is classified as non-empty if the corresponding bounding box contains at least $150$ points.

For object identification, CLIP feature-based retrieval~\cite{radford2021learning} is employed to match each buffered object to one of the $100$ scanned real-world objects in the source dataset.
Prior to deployment, the object properties of each source object are precomputed using the object property recognition scheme, which requires 31.9 seconds per object.
Each source object is represented by multiple real-world RGB patch templates covering different views (\eg, front and back) and product variants with the same geometry.
The patch templates are zero-padded to a resolution of $360 \times 360$, and their CLIP features are precomputed.
During deployment, a real-world RGB patch is captured for each non-empty buffer region and zero-padded to $360 \times 360$.
Its CLIP feature is extracted and compared with the precomputed template features using cosine similarity.
The maximum similarity across all patch templates is used as the similarity score for each source object, and the object with the highest score is selected as the identified object.
The 3D CAD model and precomputed properties of the identified object are then retrieved for subsequent planning without repeating the property recognition process.
The mean runtime of object identification is $0.03$ seconds for a five-object buffer.

Ablation studies are conducted to assess the contribution of each design choice in the object identification module.
The CLIP encoder is replaced with MobileNet-V2~\cite{sandler2018mobilenetv2} or ResNet~\cite{he2016deep}.
The multiple-template setting is replaced with a single front-view patch template per object.
The real-world patch templates are replaced with rendered patch templates generated from the corresponding virtual models of the objects in the source dataset.
Results of the ablation studies are shown in Table~\ref{tab:object_identification_ablation}.
Identification accuracy decreases when CLIP is replaced by MobileNet-V2~\cite{sandler2018mobilenetv2} or ResNet~\cite{he2016deep}, when a single patch template is used, or when captured templates are replaced by rendered templates.
These results demonstrate the contributions of the CLIP encoder, multiple templates, and captured templates to identification accuracy.
In addition, two alternative transformer-based visual encoders are evaluated.
The ViT-B/32~\cite{alex2020vit} and the Swin Transformer~\cite{liu2021swin} achieve accuracies of $96.66\%$ and $96.11\%$, respectively, indicating that the object identification module is compatible with various visual encoders.

For candidate-object selection, a virtual point cloud is sampled from the retrieved 3D CAD model of each identified object and used as input to the object Q-predictor.
The object Q-predictor requires $0.02$ seconds to predict the Q-values for all candidate objects.
The candidate object with the highest Q-value is selected as the next object for placement.

\begin{table}[t]
    \centering
    \caption{Ablation Studies on the Object Identification Design. The results are averaged across 360 instances sampled from the 100 real-world objects.}
    \label{tab:object_identification_ablation}
    \begin{tabular}{c|c}
        \hline
        &  \textbf{Accuracy}\\
        \hline
        \textbf{Full Object Identification Module} & 96.38\%\\
        \hline
        \textbf{CLIP Replaced with MobileNet-V2} & 95.28\%\\
        \hline
        \textbf{CLIP Replaced with ResNet} & 92.78\%\\
        \hline
        \textbf{Without Multiple Images} & 71.11\% \\
        \hline
        \textbf{Without Real Images} & 55.00\% \\
        \hline
    \end{tabular}
    
\end{table}

\vspace{+6pt}

\noindent\textbf{Pose Estimation.} \
For the selected object, its corresponding real-world point cloud is downsampled to a voxel size of $1\,\mathrm{mm}$ and denoised by removing statistical outliers.
The processed real-world point cloud and the virtual point cloud sampled from the 3D CAD model are utilized to estimate the object's initial pose for subsequent packing planning.
Specifically, Fast Point Feature Histograms (FPFH) descriptors are extracted from both the real-world and virtual object point clouds.
The FPFH descriptors from the two point clouds are matched to establish point correspondences, which are subsequently used by a RANSAC-based feature-matching algorithm to obtain a coarse 6D pose transformation between the two point clouds.
The Iterative Closest Point (ICP) algorithm is then performed to refine the estimated pose.
The mean runtime of pose estimation is $1.2$ seconds per object.

\vspace{+5pt}

\noindent\textbf{Placement Determination.} \
Given the initial pose of the selected object, candidate placements are generated using eight in-plane orientations about the $z$-axis, as specified in the ``Hardware and Network Settings'' subsection of Section~\ref{sec:platform}, and the convex-vertex-based locally optimal position generation procedure described in Section~\ref{sec:opa_net}.
Pose features are extracted from the candidate placements and used as input to the placement Q-predictor to determine the placement pose.
The mean runtime is $0.41$ seconds for candidate-placement generation per object and $0.02$ seconds for placement Q-predictor inference.

\vspace{+5pt}

\noindent\textbf{Pick-and-Place Action Generation.} \
For each object, the horizontal picking location is determined by transforming the geometric center of the corresponding 3D CAD model to the real-world point cloud according to the initial object pose.
The horizontal picking location is vertically projected onto the real-world object point cloud, and points within a horizontal distance of $1\,\mathrm{cm}$ from the projected location are retained as candidate surface points.
The curvature at each candidate surface point is computed, and the point with the minimum curvature is selected as the picking center $(x,y,z)$.
The approach direction for picking is defined by the surface normal at the picking center.
The placement location is predicted by the placement Q-predictor, with the vertical direction specified as the approach direction for placement.

\vspace{+5pt}
\noindent\textbf{Error Recovery.} \
The system incorporates an error recovery mechanism for handling sensing and picking failures through re-scanning and re-picking. 
After scanning the scene, the system compares the number of points in the acquired point cloud with a reference point count obtained when the scene is fully visible.
If the point count is lower than $80\%$ of the reference point count, the camera re-scans the scene.
This mechanism addresses potential sensing errors arising from the parallel execution of camera acquisition and robotic control.
A synchronization failure may cause image acquisition to occur while the robot arm remains within the camera view.
In this case, occlusion caused by the robot arm may result in missing regions in the acquired point cloud.
The re-scanning mechanism reduces the risk of planning and executing actions based on incomplete visual observations. 
The system also detects picking failures by continuously monitoring the vacuum pressure in the suction cup.
A lack of vacuum pressure reduction in the suction cup after a picking attempt indicates a picking failure.
In this case, the system terminates the packing action, retracts the robot arm from the packing area, re-scans the buffer area, and re-plans the subsequent packing action.

\subsection{Real-world Packing Results}
Real-world packing experiments are conducted using $70$ random packing sequences, each containing $10$ objects sampled from $100$ scanned real-world objects.
The proposed method is compared with four methods: (i) IR-BPP~\cite{zhao2023learning}, a reinforcement-learning-based method used as the baseline for OPA-Pack; and three packing heuristics, including (ii) Deepest Bottom-Left (DBL)~\cite{karabulut2004hybrid}, (iii) Min-Z (MINZ), and (iv) Heightmap-Minimization (HM)~\cite{wang2019stable, wang2021dense}.
All methods are evaluated using the same packing sequences.
Three metrics are used to assess packing performance: (i) compactness, defined as the ratio of the total volume of the packed objects to the container volume; (ii) the number of incompatible object pairs placed in adjacent positions with a surface distance below $\tau=5\,cm$ (\# Close Incomp. Pairs); and (iii) the number of fragile objects considered compressed when at least one heavier object is placed above them (\# Compressed Fragile) in each sequence, as determined by manual inspection.
Higher compactness values indicate higher space utilization, whereas lower values of the other two metrics indicate better performance.
Table~\ref{tab:real_system} reports the average results.

\begin{table}[t] 
  \centering
  \begin{minipage}{\linewidth}
  \caption{Comparing the average performance of OPA-Pack and the baseline over 70 real-world packing cases.
  OPA-Pack achieves compactness similar to that of the baseline while reducing the number of incompatible object pairs packed in adjacent positions and the number of compressed fragile objects.
  }
  \label{tab:real_system}
  \resizebox{\linewidth}{!}{
  \begin{tabular}{@{\hspace*{0mm}}l@{\hspace*{0.75mm}}|@{\hspace*{0.75mm}}c@{\hspace*{1.25mm}}c@{\hspace*{1.25mm}}c@{\hspace*{0mm}}}
    \hline
    \textbf{Method}
    & \textbf{Compactness $\uparrow$} & \textbf{ \# Close Incomp. Pairs $\downarrow$} & \textbf{\# Compressed Fragile $\downarrow$ } \\
    \hline
    \textbf{OPA-Pack} & \textbf{0.32} & \textbf{0.52} & \textbf{0.18} \\
    \textbf{IR-BPP (Baseline)} & \textbf{0.32} & 0.94 & 0.68 \\  
    \hline    
    \textbf{MINZ} & 0.28 & 1.26 & 0.73 \\
    \textbf{HM} & 0.29 & 1.05 & 0.88 \\
    \textbf{DBL} & 0.28 & 1.45 & 0.74 \\
    \hline
    \end{tabular}
    }
\end{minipage}
\end{table}

The proposed method and IR-BPP both achieve a packing compactness of 0.32, which is 0.04 higher than that of MINZ (0.28) and DBL (0.28) and 0.03 higher than that of HM (0.29).
Compared with IR-BPP, the proposed method reduces the average number of incompatible object pairs placed in adjacent positions from 0.94 to 0.52 and decreases the average number of compressed fragile objects from 0.68 to 0.18.

Figure~\ref{fig:real_vis} presents two real-world packing cases for the baseline and the proposed method.
Figure~\ref{fig:real_vis} (left) shows that the baseline places beverage containers above fragile fruits, including persimmons and plums, resulting in visible damage. 
In contrast, the proposed method reduces the load exerted on fragile objects by other objects placed above them.
Figure~\ref{fig:real_vis}~(right) shows that the proposed method separates incompatible object pairs, such as medical test kits and persimmons, or cockroach insecticide and persimmons, whereas the baseline places these objects in adjacent positions.
These results indicate that OPA-Pack maintains packing compactness while reducing the number of compressed fragile objects and separating incompatible object pairs.
Additional real-world packing results are provided in the supplementary material.

\subsection{Failure Case Analysis}
\begin{figure*}
    \centering
    \includegraphics[width=0.92\linewidth]{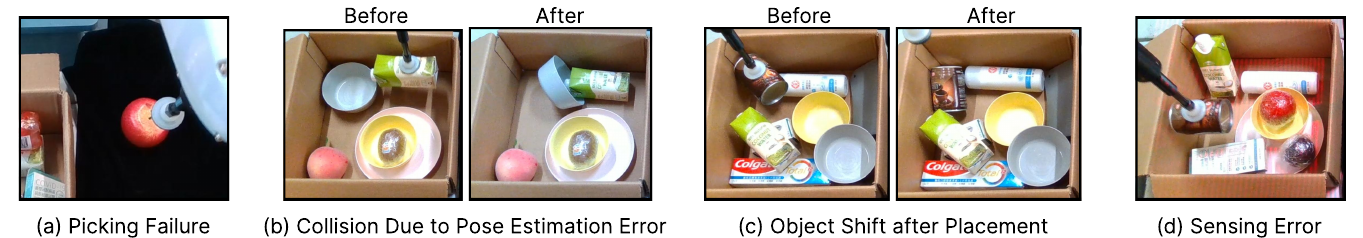}
    \caption{Visualization of representative failure cases. (a) Picking Failure: Detected by monitoring the vacuum pressure in the suction cup; the system re-scans the scene. (b) Collision due to Pose Estimation Error: Pose estimation errors may lead to collisions, impacting packing compactness. (c) Object Shift after Placement: Objects may shift post-placement; the system re-scans the scene and updates the observed container state. (d) Sensing Error: Asynchronous communication between separate threads for camera and robot controls may lead to premature scene sensing; the system detects such sensing errors and re-scans the scene.
    }
    \vspace{-4pt}
    \label{fig:real_error}
\end{figure*}

Figure~\ref{fig:real_error} presents representative failure cases in four categories:
\textbf{(a) Picking Failure:} the picking of an object may fail due to errors in picking pose generation.
This failure occurred in 2\% of the evaluated picking attempts.
The system detects the picking failure by monitoring the vacuum pressure in the suction cup, after which it re-scans the scene and repeats the picking procedure.
\textbf{(b) Collision due to Pose Estimation Error:} pose estimation errors may cause object collisions, resulting in execution failure.
This issue occurred in 4\% of the evaluated real-world trials.
\textbf{(c) Object Shift after Placement:} objects may shift after being packed in the container.
In the evaluated trials, 11\% of the placed objects exhibited a horizontal shift larger than $2\,\mathrm{cm}$.
The system re-scans the scene to obtain the updated container state for subsequent planning, thereby mitigating the impact of object shifts on subsequent packing steps.
\textbf{(d) Sensing Error:} since the camera and robot arm are controlled by separate threads, asynchronous communication may cause the camera to capture the scene while the robot arm remains within the workspace.
This issue occurred in 2\% of the evaluated real-world trials, and the system performs an additional scene scan after detecting such cases.
In the real-world experiments, 90\% of the objects were successfully processed by all three modules: object identification, object pose estimation, and picking pose generation.
As the performance of these modules can be affected by variations in real-world visual observations, improving robustness to diverse visual conditions remains an important direction for future work.

\section*{Conclusion}

This paper presents OPA-Pack, a framework that incorporates object-property information into robotic bin packing.
An object property recognition scheme based on retrieval-augmented generation and chain-of-thought reasoning is developed to predict object-centric properties.
The proposed scheme achieves an average object-property recognition accuracy above 90\%, including on objects unseen in the auxiliary database.
The predicted object-centric properties are used to derive avoidance relations between object pairs.
A dataset of 1,032 everyday objects with property annotations is constructed for learning object-property-aware packing policies.
OPA-Net is formulated to reduce load on fragile objects and separate incompatible object pairs, while maintaining compact packing.
OPA-Pack is evaluated in both a virtual simulator and on a physical platform.
In the virtual experiments, the proposed method improves avoidance accuracy from 52.3\% to 95.0\% and reduces load on fragile objects by 29.4\%, and achieves a packing compactness of 0.413, whereas the baseline achieves 0.420.
Physical-platform experiments further demonstrate a 44.68\% reduction in the average number of closely packed incompatible object pairs and a decrease in the average number of compressed fragile objects from 0.68 to 0.18.

Several limitations remain that motivate future research.
First, the proposed method primarily optimizes for predefined packing preferences, including reducing load on fragile objects and separating incompatible object pairs.
The method is evaluated on a fixed object set and does not support test-time modification of user-specified packing preferences.
Future work may investigate test-time adaptation to user-specified packing preferences for open-set objects.
Second, OPA-Pack uses sparse and discrete penalty terms in learning the object-property-aware packing policy.
Future work may investigate dense and continuous penalty functions to represent different degrees of preference violations.
Third, although the proposed method encourages compact placements, objects may still shift after placement.
Future work may incorporate object tracking to monitor packed-object locations.
Furthermore, compactness optimization does not inherently conflict with satisfying object-property constraints when the container is not full and multiple feasible placements are available.
In more constrained scenarios, such as near-full containers, the reward weights can be adjusted to prioritize packing compactness or object-property constraints according to the desired trade-off.
Future work could investigate the use of VLMs to adapt these reward weights according to task-specific packing preferences.

{\small
\bibliographystyle{IEEEtran}
\bibliography{egbib}
}

\section{Biography Section}

\vspace{-35pt}

\begin{IEEEbiography}
[{\includegraphics[width=1in,height=1.25in,clip,keepaspectratio]{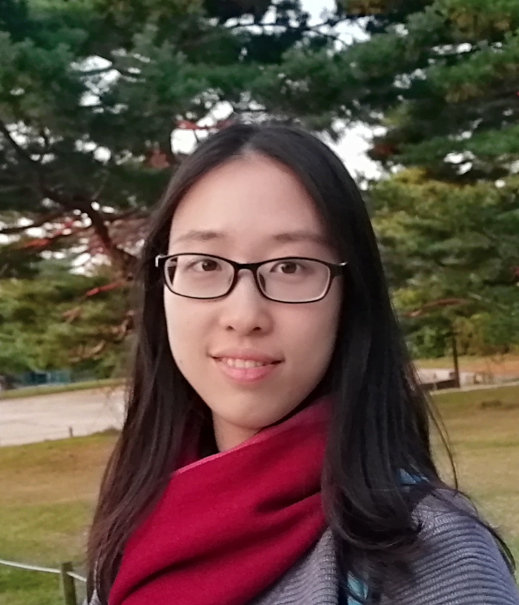}}]{Jia-Hui Pan} received the B.Sc. degree and the M.Phil degree at the School of Data and Computer Science from Sun Yat-Sen University, and the Ph.D degree in the Department of Computer Science and Engineering from the Chinese University of Hong Kong. Her research interests include robotic manipulation, 3D vision, and human action analysis.
\end{IEEEbiography}

\vspace{-30pt}

\begin{IEEEbiography}
[{\includegraphics[width=1in,height=1.25in,clip,keepaspectratio]{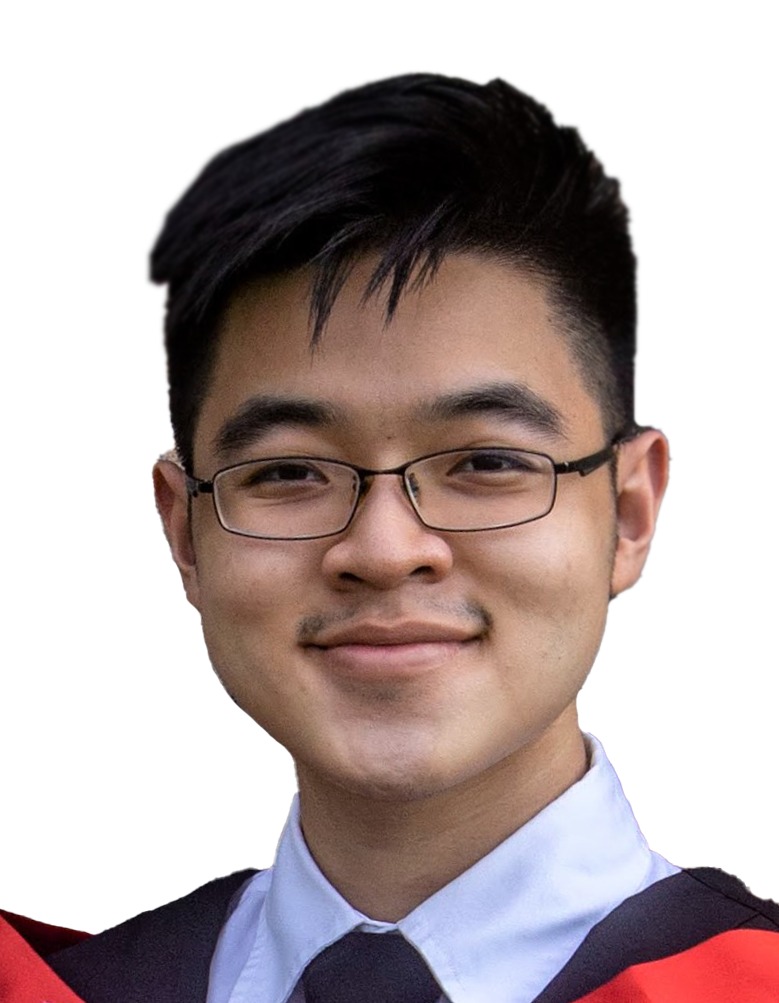}}]{Yeok Tatt Cheah} received his B.Sc. degree in Electronic and Information Engineering from The Hong Kong Polytechnic University and joined Hong Kong Centre For Logistics Robotics (HKCLR) in 2022. He is currently studying MSc in Machine Learning at University College London. His research interests include generalizable robotics, computer vision and language models. 
\end{IEEEbiography}

\vspace{-30pt}

\begin{IEEEbiography}
[{\includegraphics[width=1in,height=1.25in,clip,keepaspectratio]{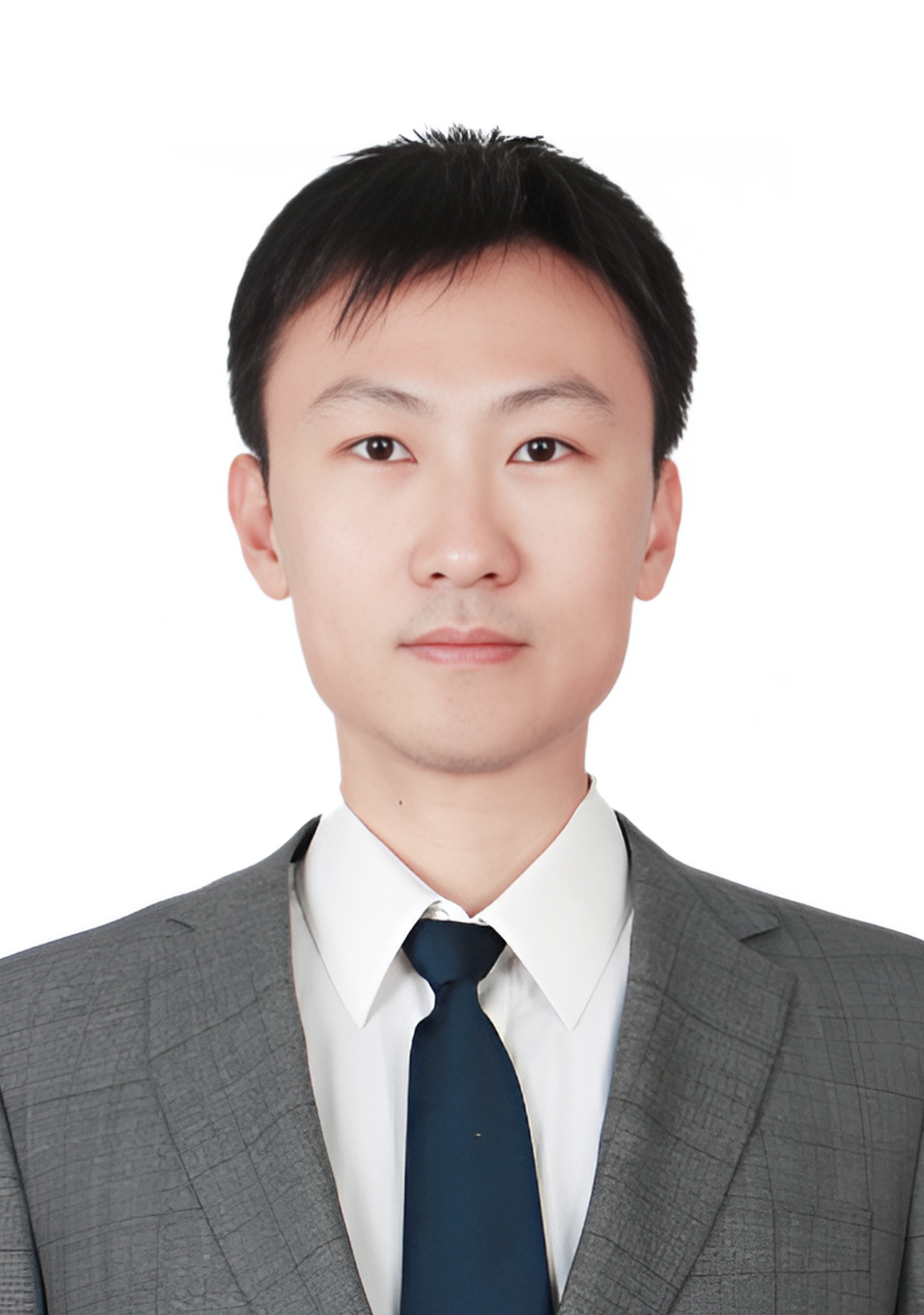}}]{Zhengzhe Liu} received his B.Sc. degree in Information Engineering at Shanghai Jiao Tong University, the M.Phil. degree in the Computer Science and Engineering Department from The Chinese University of Hong Kong, and the Ph.D degree in the Computer Science and Engineering Department from The Chinese University of Hong Kong. He worked as a postdoctoral associate at Carnegie Mellon University.
He is currently an assistant professor in the School of Data Science at Lingnan University. His research interests include 3D shape generation, 3D scene understanding, and computer graphics.
\end{IEEEbiography}

\vspace{-15pt}

\begin{IEEEbiography}
[{\includegraphics[width=1in,height=1.25in,clip,keepaspectratio]{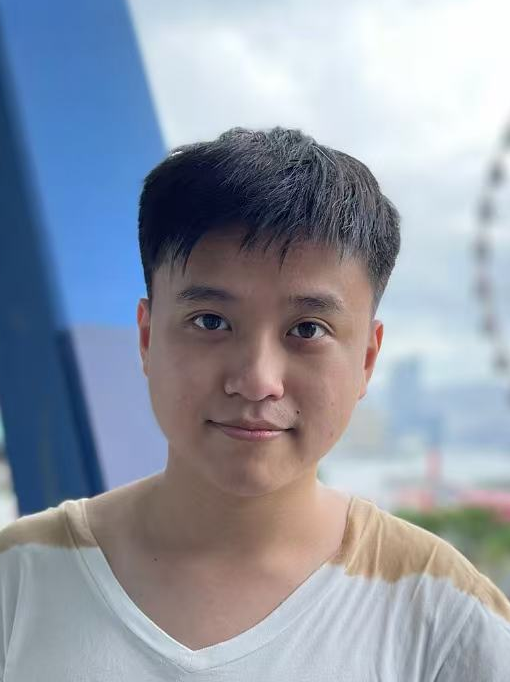}}]{Ka-Hei Hui} received the B.Sc. degree and the Ph.D. degree in the Computer Science and Engineering Department from The Chinese University of Hong Kong. He is currently a senior research scientist in Autodesk AI Lab (Canada). His research interests include 3D shape analysis, 3D generation and computer graphics.
\end{IEEEbiography}

\vspace{-15pt}

\begin{IEEEbiography}
[{\includegraphics[width=1in,height=1.25in,clip,keepaspectratio]{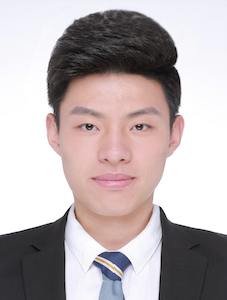}}]{Xiaojie Gao} received the Ph.D degree in the Computer Science and Engineering Department from The Chinese University of Hong Kong. He is currently a postdoctoral researcher at The Chinese University of Hong Kong. His research interests include robotic manipulation and video analysis.
\end{IEEEbiography}

\vspace{-15pt}

\begin{IEEEbiography}
[{\includegraphics[width=1in,height=1.25in,clip,keepaspectratio]{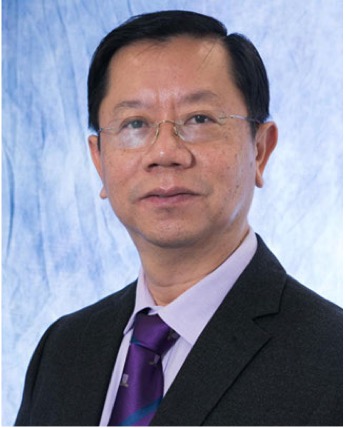}}]{Pheng Ann Heng} is Choh-Ming Li Professor of Computer Science and Engineering at The Chinese University of Hong Kong. He is the Director of the Institute of Medical Intelligence and XR. His research interests include AI/XR for medical and scientific applications, visualization, graphics, human-computer interaction and computer vision. He is recognized as a Highly Cited Researcher by Clarivate since 2024 and also recognized with the Research.com Computer Science in China Leader Award since 2023. According to Google Scholar, his publications have been cited over 80,000 times, with an H-index of 134.
\end{IEEEbiography}

\vspace{-15pt}

\begin{IEEEbiography}
[{\includegraphics[width=1in,height=1.25in, clip, keepaspectratio]{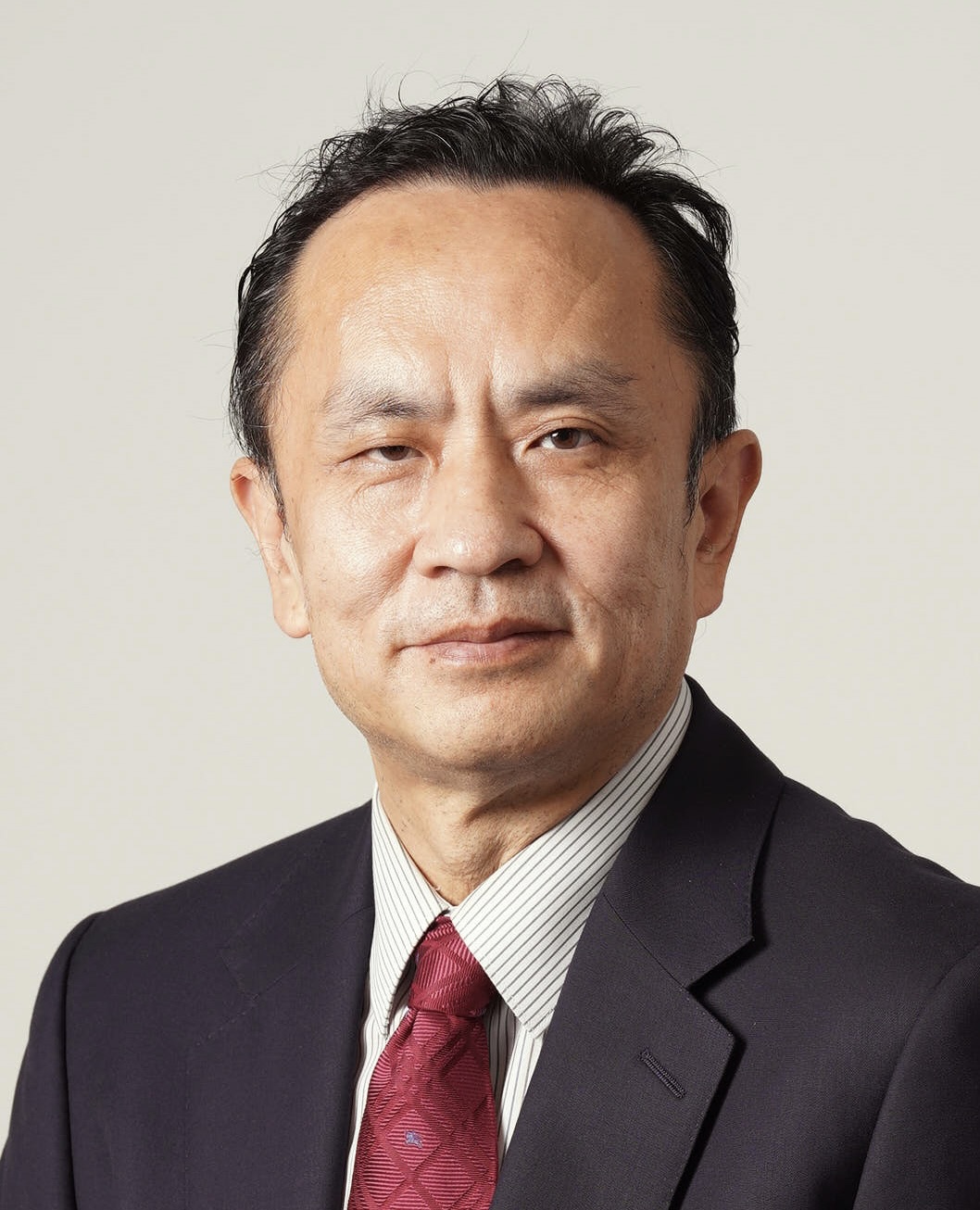}}]{Yun-Hui Liu} received the B.Eng. degree in applied dynamics from Beijing Institute of Technology, Beijing, China, in 1985, the M.Eng. degree in mechanical engineering from Osaka University, Osaka, Japan, in 1989, and the Ph.D. degree in mathematical engineering and information physics from The University of Tokyo, Tokyo, Japan, in 1992. He is currently a Professor with the Department of Mechanical and Automation Engineering and the Director of the CUHK T Stone Robotics Institute. He is also Director of Hong Kong Centre for Logistics Robotics, which is funded by the InnoHK clusters. His interests include visual servoing, medical robotics, multi-fingered robot hands, mobile robots, sensor networks, and machine intelligence.
\end{IEEEbiography}

\vspace{-15pt}

\begin{IEEEbiography}
[{\includegraphics[width=1in,height=1.25in,clip,keepaspectratio]{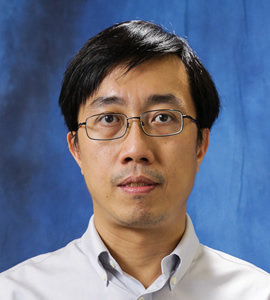}}]{Chi-Wing Fu} received the PhD degree in computer science from Indiana University Bloomington. He is currently a full professor at the Chinese University of Hong Kong. He is now serving as the Associate Editor-in-Chief (regular submissions) of IEEE Computer Graphics and Applications. He served as the program co-chair of SIGGRAPH ASIA 2016 technical brief and poster, associate editor of Computer Graphics Forum and IEEE Computer Graphics and Applications, and program committee member in various conferences such as SIGGRAPH, IEEE Visualization, and IEEE VR. His research interests include computer graphics, 3D vision, user interaction, and visualization.
\end{IEEEbiography}
\vfill

\includepdf[pages=-]{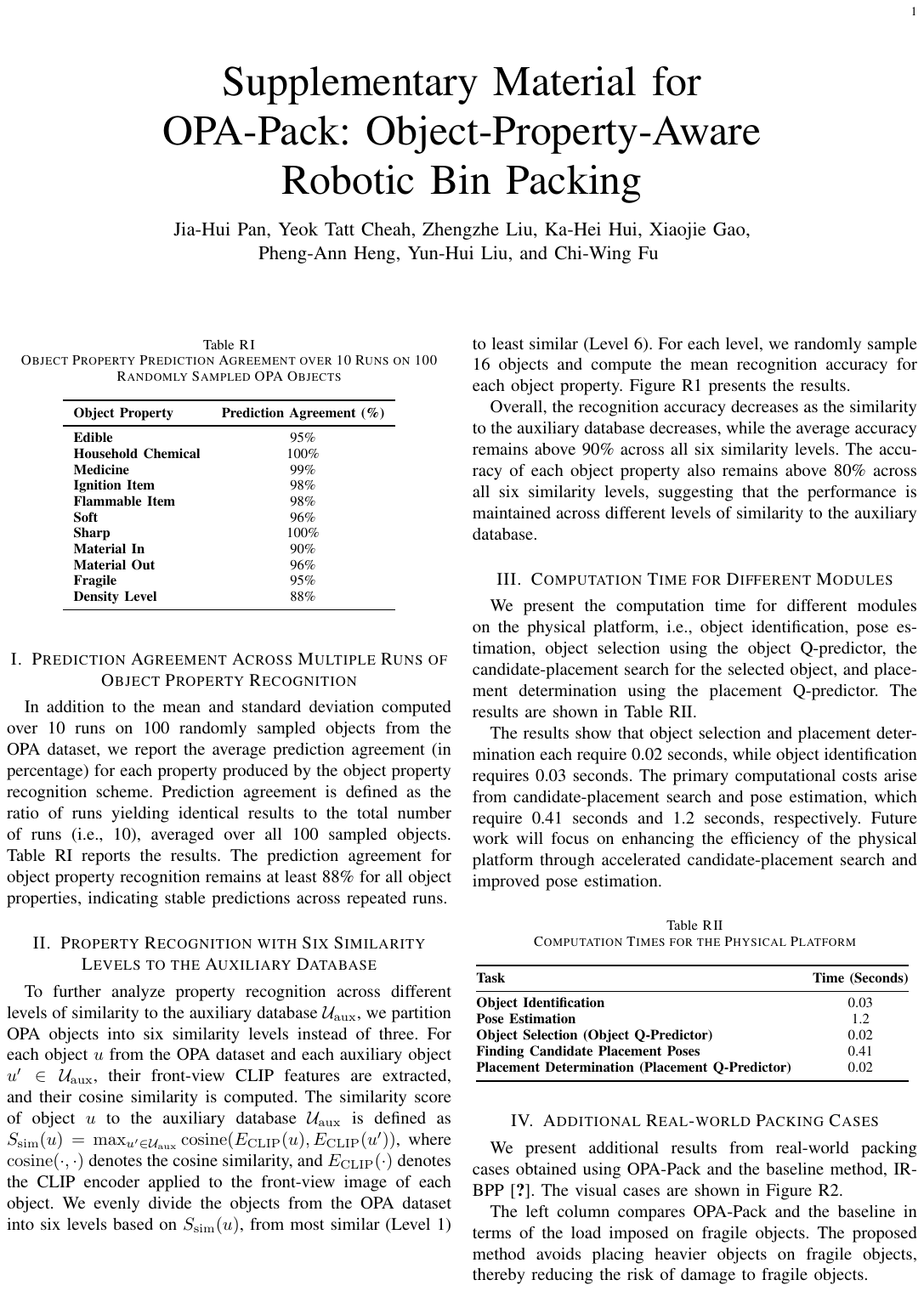}

\end{document}